\documentclass[10pt,twocolumn,letterpaper]{article}

\usepackage[pagenumbers]{cvpr} 

\definecolor{cvprblue}{rgb}{0.21,0.49,0.74}
\usepackage[pagebackref,breaklinks,colorlinks,allcolors=cvprblue]{hyperref}

\usepackage{subcaption}

\usepackage{colortbl}
\usepackage{gensymb}
\usepackage{multirow}
\usepackage{wrapfig,lipsum,booktabs}
\usepackage{pifont}
\colorlet{NiceGreen}{-red!75!green!50!blue}
\colorlet{BadRed}{red!75!green!50!blue}
\newcommand{\cmark}{\ding{51}}%
\newcommand{\xmark}{\ding{55}}%
\newcommand{\tildemid}{\raisebox{0.5ex}{\texttildelow}}
\newcommand{\best}[1]{\colorbox{NiceGreen}{\bfseries#1}}
\newcommand\hfilll{\hspace{0pt plus 1filll}}
\def\paperID{13862} 
\def\confName{CVPR}
\def\confYear{2025}

\usepackage{bibunits}
\defaultbibliographystyle{ieeenat_fullname}
\defaultbibliography{main}

\title{SphereUFormer: A U-Shaped Transformer for Spherical 360 Perception}

\author{Yaniv Benny\\
Tel Aviv University\\
\and
Lior Wolf\\
Tel Aviv University\\
}

\begin{document}
\maketitle

\begin{abstract}
    This paper proposes a novel method for omnidirectional 360$\degree$ perception.
    Most common previous methods relied on equirectangular projection. This representation is easily applicable to 2D operation layers but introduces distortions into the image.
    Other methods attempted to remove the distortions by maintaining a sphere representation but relied on complicated convolution kernels that failed to show competitive results.
    In this work, we introduce a transformer-based architecture that, by incorporating a novel ``Spherical Local Self-Attention'' and other spherically-oriented modules, successfully operates in the spherical domain and outperforms the state-of-the-art in 360$\degree$ perception benchmarks for depth estimation and semantic segmentation. 

\end{abstract}

\section{Introduction} \label{sec:intro}

Monocular omnidirectional 360$\degree$ perception is an important setting, as it allows for a full field-of-view (FOV) receptive field for the underlying model. Consequently, an entire room layout can be inferred, for example.
In contrast to regular images, a 360$\degree$ image has no boundaries. Instead, the image is horizontally cyclical, and converges to singular points at the vertical extreme points. This suggests that this domain requires a different treatment, and that best practices that work well with regular images might not have the same effect on 360$\degree$ images. To tackle this task, multiple topics need to be considered. 

\textit{How should the data be represented? What kind of distortions and irregularities are created by this representation? How will a model process this representation? How to deal with these distortions and irregularities?}

These questions have attracted researchers to construct multiple datasets~\cite{structured3d, armeni2017joint, Matterport3D, Pano3D} where depth estimation and semantic segmentation are the main tasks, and to formulate many solutions~\cite{panoformer, panocmx, egformer, unifuse, bifuse, bifuse++, hohonet, omnifusion, joint360, slicenet, 360monodepth, zhang2022bending}, which propose novel approaches to solving these tasks.
A common pattern in previous works is that they represent the data by projecting it to a 2D plane in different ways, usually equirectangular projection~\cite{miller1949equi}, cube mapping~\cite{lambers2020survey}, or patch cropping.  
Each approach has its advantages and disadvantages, but the common motivation is to represent the image in a grid formation. This enables grid architectures, such as Convolutional Neural Networks or Vision Transformers to easily be applied. This comes at a cost of distortions, irregularities, or limited FOV, that are caused by the selected projection.
A less common alternative representation that has been proposed is to represent the image in its original sphere domain by discretizing the sphere surface in some manner~\cite{flavell2010uv,keinert2015spherical,dimitrijevic2016comparison,busemann2012geometry}. The motivation is to maintain an undistorted representation of the 360$\degree$ image. However, this representation formed architectural challenges that require novel solutions. Some previous works~\cite{spheredepth,lee2019spherephd,coors2018spherenet,zhang2019orientation, zhang2020spherical, carlsson2024heal} addressed these challenges in various manners, but they fail to compete with current state-of-the-art solutions that follow a grid representation.

In this work, we tackle the drawbacks of spherical representations. 
Specifically, we introduce SphereUFormer, which is a novel U-Shaped Transformer architecture tailored for spherical 360$\degree$ perception. By directly operating on a spherical domain without distortion-inducing projections, SphereUFormer addresses the core challenges of omnidirectional perception. 
Our approach builds on the strengths of transformer models, incorporating modifications that allow it to efficiently process and understand spherical data. This includes the development of spherical-specific operations such as localized self-attention mechanism that respects the geometry of the sphere, innovative up/downsampling techniques that maintain the integrity of spherical data throughout the model, and positional encoding scheme that matches the domain properties.
SphereUFormer outperforms the state of the art in the field of 360$\degree$ perception, providing a robust and efficient tool for a wide range of applications that require comprehensive understanding of spherical environments.
In \cref{sec:representation} we describe this representation in detail. We then further describe a new method, which utilizes this representation and performs 360$\degree$ perception on spherical representations. In \cref{sec:method} we describe this method and in \cref{sec:experiments} analyze its results compared to the state-of-the-art previous works.

\section{Related Work} \label{sec:related}

\paragraph{\bf Omnidirectional perception.} There have been many advances, novelties, and research verticals in omnidirectional perception over the years. Initial works attempted standard convolutional networks on an equirectangular projection~\cite{omnidepth} (ERP). Although promising, the results were affected by the distortions created by the projection. To better handle these distortions, later works suggested technical adjustments to the convolution kernel such as Spherical Convolutions~\cite{su2017learning} or Distortion-Aware Convolutions~\cite{tateno2018distortion, coors2018spherenet}. The latter sample the convolution features from a tangent plane.
Other solutions attempted to eliminate distortions by sampling the image into smaller FOV patches~\cite{joint360, omnifusion, 360monodepth, eder2020tangent} followed by a post-processing stage that fuses the independent results. This, however, is at a cost of a restricted receptive field per patch and possible cutoff of crucial information, which either affects results or requires high overlap between patches.
Due to the advantages and limitations of each method, \cite{bifuse, bifuse++, ai2023hrdfuse} suggested a combined approach that fuses the ERP path with the individual patches at some stage of the network.
Alternative approaches avoided the use of projection, and instead, operated directly on the sphere by constructing a spherical mesh. Usually either an icosphere~\cite{spheredepth, deepsphere, jiang2019spherical, lee2019spherephd} or healpix~\cite{deepsphere, carlsson2024heal}, and apply Graph Convolution~\cite{zhang2019graph}, Mesh Convolution~\cite{hu2022subdivision}, or some custom convolution-like operation.

\paragraph{\bf Vision Transformers.} ViT's~\cite{dosovitskiy2020image} have disrupted the field of computer vision, from general-purpose architectures~\cite{ranftl2021vision, wang2022uformer} to specific purposes~\cite{carion2020end, cheng2021per, xie2021segformer}. The attention mechanism~\cite{bahdanau2014neural, luong2015effective, vaswani2017attention} allows for dynamic pooling of features based on relevance and structural position. ViT's benefit from an efficient large context window compared to their CNN counterparts, which allows them to more easily learn long-distance relations between features. While earlier omnidirectional perception works proposed solutions based on CNN's, recent contributions replaced the convolution kernels with local attention of various forms~\cite{panoformer, egformer, panocmx} with favorable results. This work follows this direction. It utilizes the foundation of UFormer~\cite{wang2022uformer} and adapts it to spherical representations with custom attention layers.

\paragraph{\bf Graph Attention.}  Graph Convolution Networks~\cite{zhang2019graph} (GCN) are neural networks that apply convolution-like operations on data represented as a graph. A graph convolution operation aggregates the results of local operations between a node and each of its neighbors. Over multiple layers, this allows for information to pass between distant nodes in the graph and the update of these nodes. Graph Attention Networks~\cite{velivckovic2017graph} were proposed to enhance Graph Convolution Networks, thereby improving the message-passing capabilities of the GCN. The attention mechanism has the benefit of allowing the operations to focus on a few sets of relevant nodes and providing an efficient method to encode positional information in the graph. For Geometric Graphs~\cite{pei2020geom} geometric location is encoded to inform of each node's absolute and relative position, thereby making nodes aware of their global position and their respective neighbors' relative position.

\section{Sphere Representation} \label{sec:representation}

\begin{figure}[t]
\centering
\begin{subfigure}{0.24\linewidth}
\includegraphics[width=\textwidth, trim={50 100 1550 100}, clip]{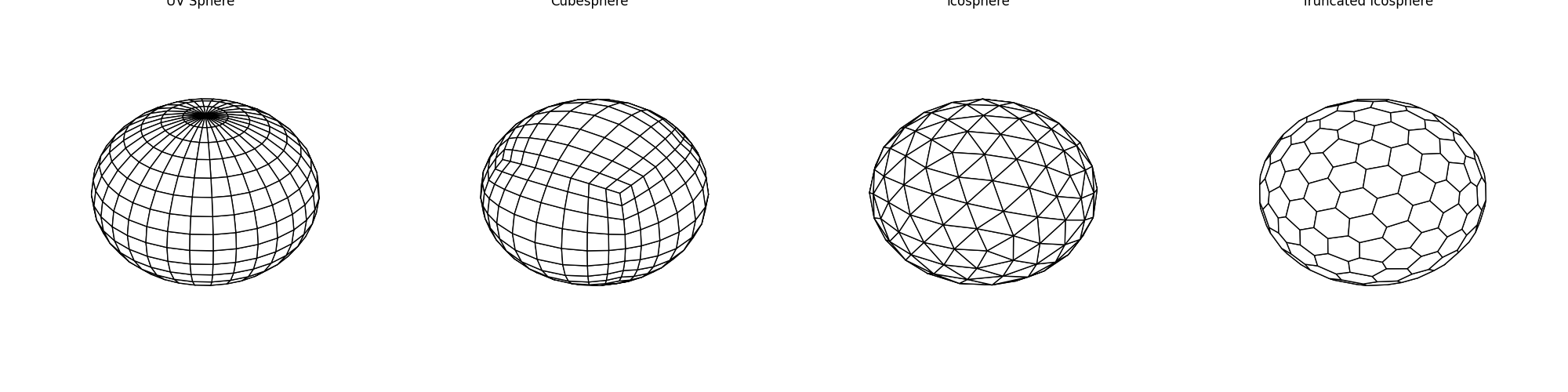}
\subcaption{}
\label{subfig:uvsphere}
\end{subfigure}
\begin{subfigure}{0.24\linewidth}
\includegraphics[width=\textwidth, trim={550 100 1050 100}, clip]{images/blank/allspheres.png}
\subcaption{}
\label{subfig:cubesphere}
\end{subfigure}
\begin{subfigure}{0.24\linewidth}
\includegraphics[width=\textwidth, trim={1050 100 550 100}, clip]{images/blank/allspheres.png}
\subcaption{}
\label{subfig:icosphere}
\end{subfigure}
\begin{subfigure}{0.24\linewidth}
\includegraphics[width=\textwidth, trim={1550 100 50 100}, clip]{images/blank/allspheres.png}
\subcaption{}
\label{subfig:hexasphere}
\end{subfigure}
\caption{{\bf Sphere Representations.} From left to right: uvsphere, cubesphere, icosphere, hexasphere.}
\label{fig:allspheres2}
\end{figure}

Arguably the first topic in question when attempting to solve a problem is how to formulate the data. For omnidirectional 360$\degree$ images, the data can be seen as an array of values $v_{\phi,\theta}$ assigned to a ray pointing from a center point-of-view towards $(\theta, \phi)$ for $\theta \in [0, 2\pi]$ being the horizontal angle and $\phi \in [0, \pi]$ the vertical. The value $v$ can correspond to the RGB value, depth, object category, etc.
By selecting to sample on finite such angle pairs, the sphere becomes discretized. The sampling method determines how the sampled data can be processed and how successful the solution will become.

The most common and straightforward is a uvsphere~\cite{flavell2010uv}, where points are sampled in a grid with equal spacing for $(\theta, \phi)$. See \cref{subfig:uvsphere}. The uvsphere has a very high degree of horizontal symmetry, which is a good property. Another advantage is that it allows an unwrapping of the sphere into a 2D matrix. This unwrapping is called equirectangular projection (ERP). The downside of this representation is that the density of samples on the sphere is higher around the poles, due to the decreasing cross-section. This creates an imbalanced effective resolution over the sphere, and the projected image in 2D is distorted.

An alternative that mitigates imbalanced resolution and distortion is cube mapping (or cubesphere)~\cite{dimitrijevic2016comparison} (\cref{subfig:cubesphere}), which projects the sphere onto the faces of a cube. Sample resolution is controlled by deciding how many rectangles each face of the cube is subdivided into. This removes the sampling density around the poles and the sphere can be unwrapped into six 2D images. However, two of these images are not vertically oriented and require special treatment. In addition, the transition between faces is not fluent, and complex padding methods~\cite{cheng2018cube} can only increase the receptive field of each image to some extent. Since the images are separated, a post-processing stage needs to fuse their predictions to form a unified prediction on the sphere. One option is to operate on the cubesphere mesh instead of the 2D projection. Unfortunately, the cubesphere has irregularities at the coordinates of the cube corners and does not have a high horizontal symmetry, which makes it less optimal for this purpose.

Instead of projecting the data to a grid and then having to deal with its limitations, a more elegant approach can be formulated using the sphere representation directly. Aside from the previously mentioned options, a common way to discretize a sphere is with Geodesic (icosphere, \cref{subfig:icosphere}) and Goldberg (hexasphere, \cref{subfig:hexasphere}) polyhedrons~\cite{busemann2012geometry}. An icosphere is created by repeatedly subdividing an icosahedron into smaller triangles. A hexasphere is the dual polyhedron where vertices are swapped with face normals, and edge connections with face adjacency. These types of sphere representation have a very high degree of symmetry and have a very evenly spaced sampling on the sphere surface. Due to their duality, our work follows the icosphere mesh, and icosphere/hexasphere mode is toggled by whether data points $(\theta, \phi)$ describe mesh vertices (hexasphere) or face normals (icosphere). Therefore we will sometimes refer to icospheres/hexaspheres by referring to the ``node type'' (face/vertex respectively).
Other notable options for sphere sampling are Fibonacci spheres~\cite{keinert2015spherical} and HealPix~\cite{gorski1998analysis}.

\begin{figure}[t]
\centering
\begin{tabular}{@{}c@{}c@{}c@{}c@{}c@{}}
\includegraphics[width=0.2\linewidth, trim={0 0 0 0}, clip]{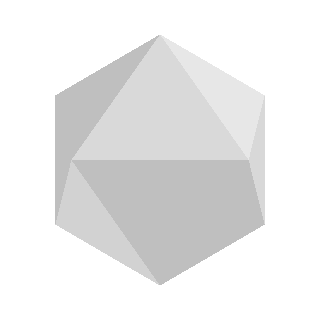} &
\includegraphics[width=0.2\linewidth, trim={0 0 0 0}, clip]{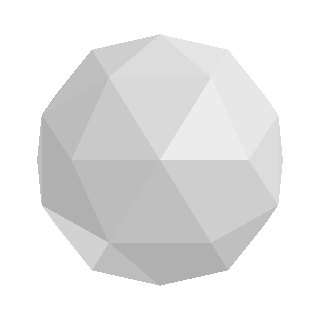} &
\includegraphics[width=0.2\linewidth, trim={0 0 0 0}, clip]{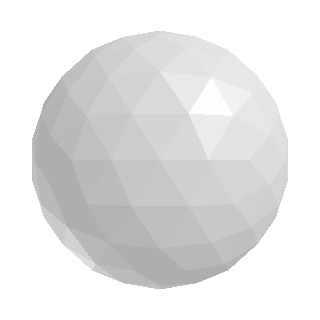} &
\includegraphics[width=0.2\linewidth, trim={0 0 0 0}, clip]{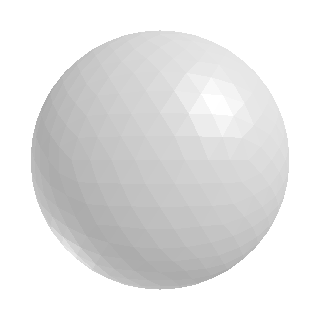} &
\includegraphics[width=0.2\linewidth, trim={0 0 0 0}, clip]{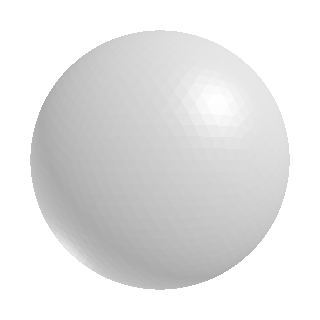} \\
0 & 1 & 2 & 3 & 4 \\
\end{tabular}
\caption{{\bf Icospheres of different ranks.} An increase in rank is made by subdividing each triangle into 4 smaller triangles, and results in an increased resolution.}
\label{fig:spheres}
\end{figure}

\begin{table}[t]
    \centering
    \begin{tabular}{l@{~}c@{~}c@{~}c@{~}c@{~}c@{~}c@{~}c@{~}c}
    \toprule
    Rank: &  0 & 1 & 2 & 3 & 4 & 5 & 6 & 7  \\
    Faces: &  20 & 80 & 320 & 1280 & 5120 & \tildemid20K & \tildemid82K & \tildemid328K  \\ 
    Vertices: & 12 & 42 & 162 & 642 & 2562 & \tildemid10K & \tildemid41K & \tildemid164K  \\ 
    \bottomrule
    \end{tabular}
    \caption{{\bf Icosphere Properties.} Number of faces $(20\times4^i)$ and number of vertices $(10\times4^i+2)$ per subdivision rank $i$.}
  \label{tab:spheres}
\end{table}

\begin{figure}[t]
\centering
\begin{subfigure}{0.15\linewidth}
\includegraphics[width=\textwidth, trim={30 0 30 0}, clip]{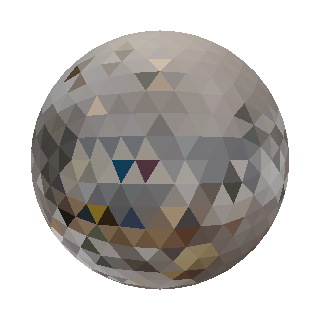}
\subcaption*{(a1)}
\end{subfigure}
\begin{subfigure}{0.15\linewidth}
\includegraphics[width=\textwidth, trim={30 0 30 0}, clip]{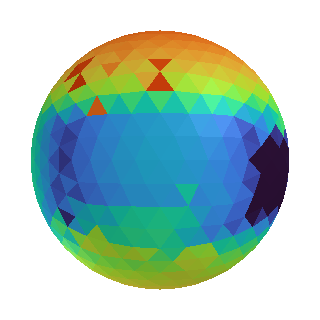}
\subcaption*{(b1)}
\end{subfigure}
\begin{subfigure}{0.15\linewidth}
\includegraphics[width=\textwidth, trim={30 0 30 0}, clip]{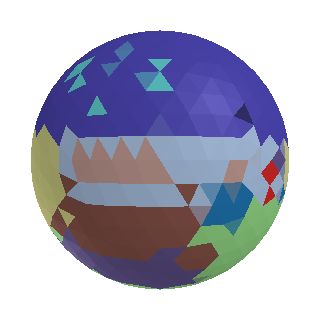}
\subcaption*{(c1)}
\end{subfigure}
\begin{subfigure}{0.15\linewidth}
\includegraphics[width=\textwidth, trim={30 0 30 0}, clip]{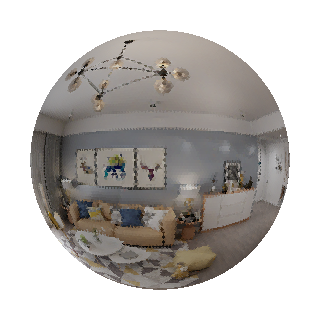}
\subcaption*{(a2)}
\end{subfigure}
\begin{subfigure}{0.15\linewidth}
\includegraphics[width=\textwidth, trim={30 0 30 0}, clip]{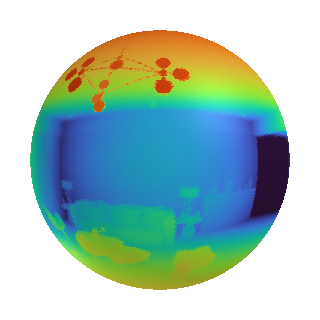}
\subcaption*{(b2)}
\end{subfigure}
\begin{subfigure}{0.15\linewidth}
\includegraphics[width=\textwidth, trim={30 0 30 0}, clip]{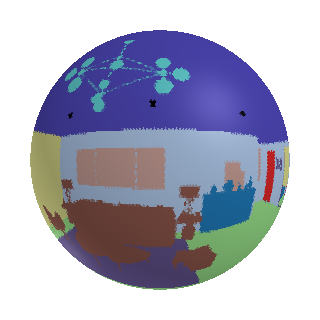}
\subcaption*{(c2)}
\end{subfigure}
\caption{{\bf Data as Icospheres.} Icosphere of rank 3 (a1-c1) and rank 6 (a2-c2). (a) RGB, (b) Depth Map, and (c) Semantic Layout.}
\label{fig:spheres_filled}
\end{figure}

Icospheres have the highest order of symmetry, the most uniform distribution of points (perfect uniform distribution is impossible), easy setup, and an inherent up/down sampling definition through subdivision. All these reasons make them ideal for discrete spherical representation. As \cref{fig:spheres} and \cref{tab:spheres} show, in different ranks, icospheres produce sphere discretization in various resolutions. Each increase in sphere rank divides each triangle face into four smaller ones, hereby gradually increasing the sphere resolution.
An example of representing data with icospheres can be seen in \cref{fig:spheres_filled}. In low resolution, the image seems ``pixelated'' as the independent triangle faces are distinguishable. But, with a high enough resolution, the image becomes detailed and visibly smooth. Our method uses high-resolution icospheres to represent the 360$\degree$ image, and through up and downsampling on the sphere structure, uses low-resolution spheres as hidden layers inside the model.

\section{Method} \label{sec:method}

\begin{figure*}[t]
\centering
\includegraphics[width=\textwidth, trim={10 377 10 262}, clip]{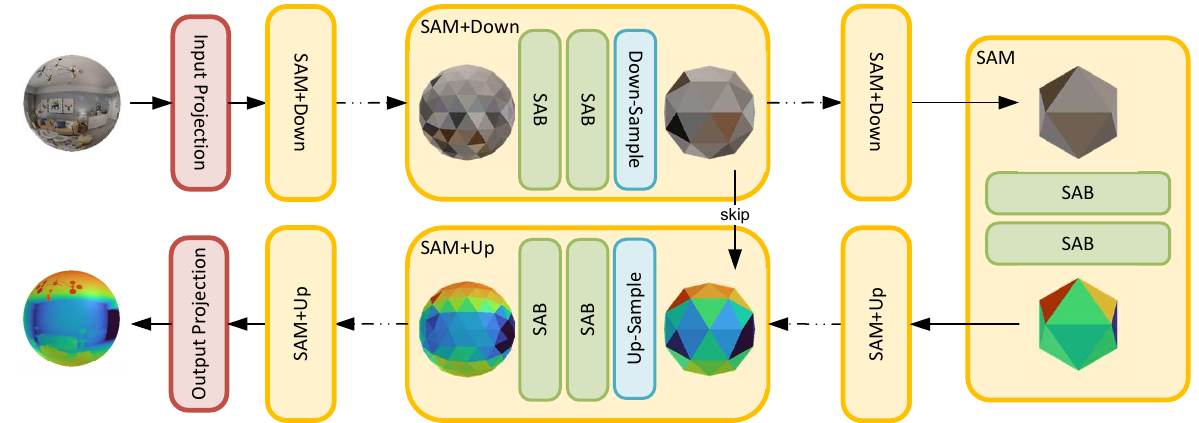}
\caption{{\bf The SphereUFormer architecture.} A spherical representation is fed into the model. A linear input projection layer encodes the RGB values to latent embedding vectors. A sequence of SAM modules apply local self-attention on the spherical data along with downsampling layers that gradually reduce the resolution of the sphere. A sequence of SAM modules along with upsampling layer and bypass skip connections decode the data. An output projection converts the latent embeddings to the output channel size.}
\label{fig:arch}
\end{figure*}

This section introduces \textbf{SphereUFormer}, our proposed U-Shaped Transformer for Spherical 360$\degree$ Perception.
As the name suggests, the model is influenced by UFormer~\cite{wang2022uformer}, with significant modifications to support spherically structured data. A high-level diagram is depicted in \cref{fig:arch}. The model's components are described in this section. For technical implementation details, see the supplementary.

The network begins with an input projection that applies linear projection from the RGB values to latent embedding vectors. The projection output is then passed into an Encoder-Decoder network with multiple levels of self-attention modules in different sphere resolutions. We call these modules ``Spherical Attention Modules'' (SAM). At the bottom of the network lies a bottleneck SAM module that operates on the lowest resolution. Skip connections pass from the encoder to the decoder and feed high-resolution features. Finally, a linear Output Projection projects the decoder output to the output dimensions.

As described in \cref{sec:representation}, our solution features the icosphere as its spherical representation structure, as we have found it to be the most intuitive and to contain the most satisfying properties. But it is not limited to it, and with minor modifications supports any spherical structure.

Our method contains many local operations on the graph, from the down and up sampling, to the self-attention mechanism. Computing the cosine similarity or finding the neighbors of a node to generate the mapping scheme is slightly costly. This would be a problem if it was performed on each iteration, as the model would be slowed down. Luckily, since the icosphere graph is fixed, these mappings only need to be computed once in advance. Therefore there is no need to recompute it in every message massing operation.

\subsection{Up/Down Sampling}
In the spherical setting, upsampling and downsampling operations are performed by transforming a spherical graph into a new one with more or fewer nodes, respectively. This can be difficult and inefficient to compute for two arbitrary spherical polyhedrons. However, icospheres have multiple elegant and desired properties. 
First, an increase in resolution by one rank is the result of splitting each triangle face into four smaller ones, with one central triangle that shares its center with the large one.
Second, for the same reason, each vertex in the low-resolution icosphere also exists in the high-resolution one, with new vertices created at the center of each edge connection.
Third, the number of nodes/faces increases at a constant rate.

All these properties make constructing up and down operations extremely straightforward and efficient. Every common pooling technique, such as max pooling, average pooling, center pooling, and interpolation, can be implemented. However, there are preferences for each setting.
For downsampling, when the data nodes are represented by the faces, we have found max and center pooling to have the same overall results. When the data nodes are on the vertices (equivalent to hexasphere), max and average pooling become a little complicated, since there are both instances of five and six neighbors in the graph. Therefore, we resorted to only center pooling.
For upsampling, the faces mode most simply operates with nearest upsampling, where each triangle receives the value according to the lower resolution triangle it was subdivided from. Other interpolation methods require more complicated computations and were therefore not considered. For vertices, however, interpolation is very simple, since each new node is located precisely at the center of an existing edge. For more upscaled upsampling, Barycenteric coordinates~\cite{floater2015generalized} interpolation can also be performed quite easily, but we did not find a reason to upscale beyond a factor of 2 at a time.

\subsection{Global Positional Encoding}
Positional encoding is a common and critical technique in Transformers~\cite{vaswani2017attention, radford2018improving} and Vision Transformers~\cite{dosovitskiy2020image, ranftl2021vision, carion2020end, cheng2021per, xie2021segformer, wang2022uformer}. SphereUFormer utilizes positional encoding as well, but in a manner that suits spherical representations. The node position on the unit sphere is expressed using spherical coordinates ($\theta, \phi$).
In our setting of omnidirectional perception, the data is always vertically aligned, therefore the model does not need to be vertically equivariant. However, the horizontal rotation is arbitrary, therefore it is desired to make the model horizontally equivariant to rotation and flipping.
We therefore apply absolute positional encoding for the vertical position but not for the horizontal position. Horizontal position is only used in relative to another node as position bias in the Local Self-Attention described in the next section. 

For the vertical position, we consider $\phi \in [0, \pi]$. The position was encoded by applying sinusoidal encoding on the position value, then passed through an MLP with an output dimension equal to the embedding dimension. We apply the vertical global positional encoding on the projected input and on the queries and keys in each attention module. 
More details can be found in the supplement.

\subsection{Spherical Local Self-Attention}

\begin{figure}[t]
\centering
\includegraphics[width=\linewidth, trim={170 602 8 29}, clip]{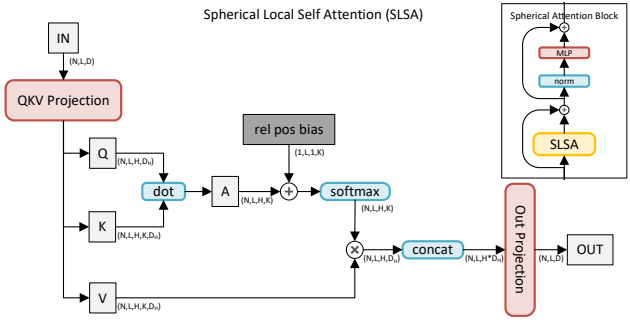}
\caption{{\bf Spherical Local Self Attention.} Attention is applied between each data node and its $K$ neighbors. A learned relative position bias encodes information about the neighbors' relative position. In the right corner is a diagram of the enclosing block.}
\label{fig:slsa}
\end{figure}

\begin{figure}
\centering
\includegraphics[width=\linewidth, trim={20 320 20 256}, clip]{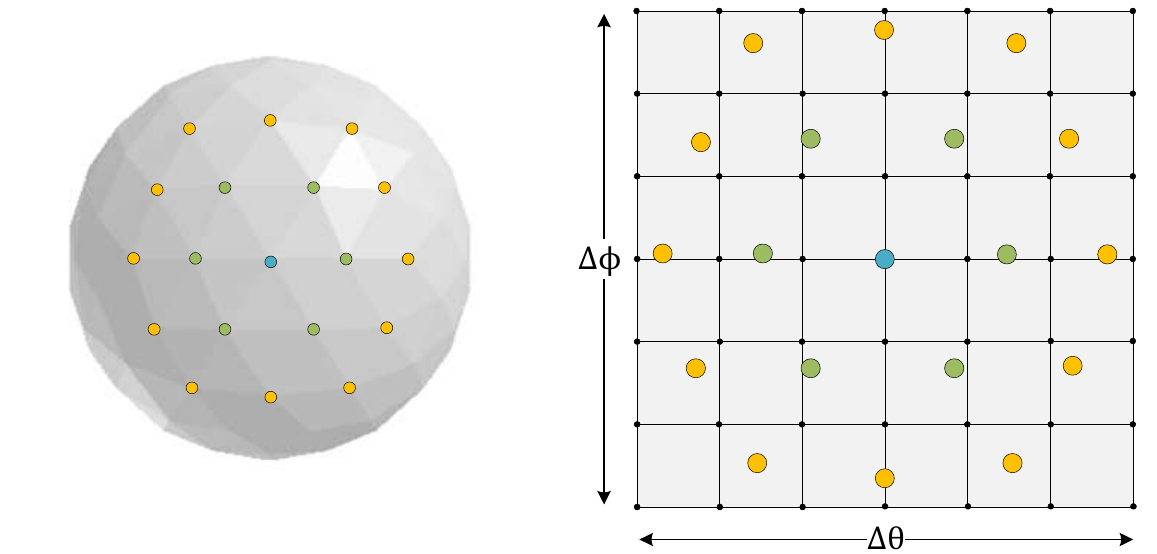}
\caption{{\bf Relative Positional Encondig.} On the left, a center point (blue) and its first (green) and second (yellow) degree neighbors are defined by the graph structure. 
The relative position kernel is defined by a 7$\times$7 grid, represented by the black nodes in the right image.
The center point is positioned at the center of the grid and the neighbors are distributed across the grid according to their relative change in horizontal ($\Delta\theta$) and vertical ($\Delta\phi$) angles.
Bilinear interpolation samples a learned positional encoding.}
\label{fig:rel_pos}
\end{figure}

The Spherical Local Self-Attention is the main component in our network that drives the SphereUFormer's 360$\degree$ perception. A detailed diagram is shown in \cref{fig:slsa}. Given an input sphere, each node $x_i$ in the sphere is projected into a query ($q_i$), key ($k_i$), and value ($v_i$). These vectors are all of the shape ($N, L, H, D_H$), where $H$ is the number of heads and $D_H$ is the head dimension. Similar to standard practices, $D_H$ is derived from the input dimension $D$ and the number of heads. However, since the self-attention mechanism is the only spatial operation in the network, we added a ``head dimension coefficient'' ($C_{head}$) hyperparameter to upscale the head dimension $D_H$ as a reverse bottleneck. Thus: $D_H = (D / H) \cdot C_{head}$. This coefficient allows an increase in the number of attention heads without reducing the dimension of each head and without increasing the overall size of the entire network, resulting in only a minor increase in parameters. We further discuss the motivation of this in the ablation section (\cref{sec:ablation}).

For each query, the neighboring $K$ vectors are grouped according to the sphere geometry. Attention for each query is performed on the selected neighboring group. The number of neighbors $K$ depends on a hyperparameter named ``window coefficient'' ($C_{win}$), which dictates the N-order neighbors to collect. A window coefficient of 0 provides no neighbors, a coefficient of 1 provides only first-order neighbors, 2 also provides second-order neighbors, and so forth. Note that due to irregularities in the graph structure, the number of keys for each node is not identical as some have slightly fewer neighbors than others. This is homogenized for parallelization purposes by adding null neighbor keys and masking them.

For relative position information, a learned relative position bias is added to the attention map. Since having a learned parameter per query-key pair is both memory expensive and harmful in terms of inductive bias, we took a shared-weights approach. For each node $i\in[1,L]$, we measure the angular difference ($\Delta\phi_{i,k}, \Delta\theta_{i,k}$) for each key $k\in[1,K]$ and normalize them by the max deltas. This is only computed once during initialization.
During runtime, the normalized deltas (in $[-1,1]$) are used to sample learned parameters from a 7$\times$7 window. This sampling from a non-linear 2D function acts as a cheap relative positional encoding for the query-key pairs. ~\cref{fig:rel_pos} illustrates the relative position implementation. Given the blue center points, its first-order neighbors are depicted in green, and second-order neighbors in yellow, are placed in their relative position. A 7$\times$7 grid, represented by the black points, represents the learned weights of the grid. Each neighbor gets its positional encoding by bilinearly interpolating between the four nearby weights.

\section{Experiments} \label{sec:experiments}

We evaluate our method by comparing it with state-of-the-art methods for depth estimation and semantic segmentation. For this purpose, we used the \textbf{Stanford2D3D}~\cite{armeni2017joint} and \textbf{Structured3D}~\cite{structured3d} datasets. Both contain RGB, depth, and semantic segmentation data.
We trained our model and all baselines using the training configuration of PanoFormer~\cite{panoformer}. For more information on the training protocol please refer to the supplementary. 

We compared against ERP methods~\cite{panocmx,panoformer,egformer}, spherical methods~\cite{zhang2019orientation,carlsson2024heal}, and Elite360D~\cite{ai2024elite360d}, that utilizes a combination of ERP with a low-resolution icosphere.
We did not compare against patch-based solutions, as the difference in compute resources makes an unfair comparison.

\subsection{Design Decisions}\label{sec:decisions}

Certain design decisions were made for the evaluation section. Some due to fairness considerations when comparing to the baselines, and others after testing various alternatives.

\begin{table}[t]
    \centering
    \begin{tabular}{l@{~~~}c@{~~}c@{~~}c}
    \toprule
    \textbf{Model}   & \textbf{Res. (\#Pixels)}     & \textbf{Params} & \textbf{Flops} \\                   
    \midrule
        PanoFormer\hfilll~\cite{panoformer}          & 256$\times$512\hfilll(131K)              & 14.5M & 11.8G \\
        EGFormer\hfilll~\cite{egformer}              & 256$\times$512\hfilll(131K)              & 15.2M & 15.6G \\
        SFSS\hfilll~\cite{panocmx}                   & 256$\times$512\hfilll(131K)              & 15.1M & 18.9G \\
        Elite360D\hfilll~\cite{ai2024elite360d}      & 256$\times$512\hfilll$^\dagger$(136K)    & 14.7M & 13.6G \\
        HealSWIN\hfilll~\cite{carlsson2024heal}      & $12\cdot4^7$\hfilll(196K)                & 12.0M & 39.0G \\
        HexRUnet\hfilll~\cite{zhang2019orientation}  &  $10\cdot4^7+2$\hfilll(164K)             & 14.0M & 12.4G \\
    \midrule
        PanoFormer\hfilll~\cite{panoformer}          & 512$\times$1024\hfilll(524K)             & 14.5M & 44.7G \\
        EGFormer\hfilll~\cite{egformer}              & 512$\times$1024\hfilll(524K)             & 15.2M & 65.8G \\
        SFSS\hfilll~\cite{panocmx}                   & 512$\times$1024\hfilll(524K)             & 15.1M & 80.3G \\
        Elite360D\hfilll~\cite{ai2024elite360d}      & 512$\times$1024\hfilll$^\dagger$(529K)   & 14.7M & 51.4G \\
        HealSWIN\hfilll~\cite{carlsson2024heal}      & $12\cdot4^8$\hfilll(786K)                & 12.0M & 156.1G \\
    \bottomrule
    \end{tabular}
    \caption{Resolution and parameter count of baseline models. $^\dagger$Elite360D uses additional icosphere of rank 4 (5120 data points).}
    \label{tab:paramstheirs}
\end{table}

\begin{table}[t]
    \centering
    \begin{tabular}{l@{~~~}c@{~}c@{~~~}c@{~}c@{~}c@{~~~}c@{~~}c}
    \toprule
    \# & \textbf{Rank} & \textbf{Type} & \textbf{$\text{C}_{head}$} & \textbf{$\text{C}_{win}$} & \textbf{Res.}     & \textbf{Params} & \textbf{Flops} \\                   
        \midrule
        & 6 & hex       & 1 & 1 & 41K & 11.2M & 2.5G \\
        & 6 & ico       & 1 & 1 & 82K & 11.2M & 4.9G \\
        & 7 & hex       & 1 & 1 & 164K & 11.2M & 9.9G \\
        & 7 & ico       & 1 & 1 & 328K & 11.2M & 19.1G \\
        & 7 & hex       & 2 & 1 & 164K & 14.9M & 13.0G \\
        \rowcolor{NiceGreen}
        (1) & \bf 7 & \bf hex  & \bf 2 & \bf 2 & \bf 164K & \bf 14.9M & \bf 13.1G \\
        \rowcolor{NiceGreen}
        (2) & \bf 8 & \bf hex  & \bf 2 & \bf 2 & \bf 655K & \bf 14.9M & \bf 52.7G \\
        
    \bottomrule
    \end{tabular}
    \caption{Resolution (total data points) and parameter count of various configurations of our model. Selected configurations in green.}
    \label{tab:paramsours}
\end{table}

For fairness reasons, it is important to compare methods of similar size. We configured all models to have 4 downsampling stages, and a depth (number of blocks per stage) of 2 for each scale. \cref{tab:paramstheirs} show the parameter and flop count per model. We gave PanoFormer~\cite{panoformer}, EGFormer~\cite{egformer}, Elite360D~\cite{ai2024elite360d} and HerRUNet\cite{zhang2019orientation} base embedding dim of 32, SFSS~\cite{panocmx} 48, HealSWIN~\cite{carlsson2024heal} 64. The resnet18 backbone was used for Elite360D. 
This resulted in as equal as possible resolution, parameter count, and flops that the different architectures allow.
We analyzed input resolutions of 256$\times$512 and 512$\times$1024 for the grid models. Resolution of the spherical models was chosen to be as close as possible to those pixel counts. This was with subdivision ranks 7 and 8 respectively. Also for fairness, none of the methods used pretrained weights. To evaluate all models fairly and on equal terms, the predictions in the various representations were projected uniformly to the sphere surface. Unlike evaluating directly on ERP, this provides an evaluation that evenly averages over the sphere surface and is not biased to extreme vertical angles.

We followed these guidelines to select the sphere setting with the closest number of nodes, total parameters and flops to the baselines. These stats are reported in \cref{tab:paramstheirs} for the existing methods, while \cref{tab:paramsours} lists variations of our method concerning: sphere rank, node type, head, and window coefficients ($C_{head}, C_{win}$). As can be seen, spheres with a rank of 6 have a very low resolution, regardless of the node type. For spheres with rank 7, we have found that the ``vertex'' node type (with 164K nodes) provides a resolution that is closest to the 256$\times$512 (total of 131K pixels) resolution, and rank 8 is the closest to 512$\times$1024. 
The base configuration of our method has fewer parameters and flops than the baselines. This is largely because, unlike most baselines, our method does not contain spatial layers, such as convolution kernels, outside of the attention layer. As a result, the base configuration of our method is less powerful. This is also further addressed in the ablation section (\cref{sec:ablation}). To minimize this difference, we have found that increasing the head coefficient results in a model with slightly more parameters to match the minimum of the baselines. Increasing the window coefficient does not add any parameters and an insignificant amount of flops. Finally, we have selected two configurations for comparison with the baselines. One with rank 7 to be compared to 256$\times$512 models, and one in rank 8 for larger resolution. These are highlighted in \cref{tab:paramsours}.

\begin{table*}[t]
    \centering
    \begin{tabular}{@{}l@{~}|@{~}c@{~~}c@{~}|@{~}c@{~}c@{~}c@{~}|@{~~}c@{~}c@{~}c@{~~}|@{~~}c@{~}c@{~~}|@{~~}c@{~}c@{}}
    \toprule
     & & & \multicolumn{6}{c}{\textbf{Depth Estimation}} & \multicolumn{4}{c}{\textbf{Semantic Segmentation}} \\
     & & & \multicolumn{3}{c}{\textbf{Stanford2D3D}}     & \multicolumn{3}{c}{\textbf{Structured3D}} & \multicolumn{2}{c}{\textbf{Stanford2D3D}}     & \multicolumn{2}{c}{\textbf{Structured3D}} \\
    \textbf{Model} & \textbf{\#Params} & \textbf{\#Flops} &  \textbf{MAE$\downarrow$} & \textbf{MRE$\downarrow$} & \textbf{$\delta_1$$\uparrow$} &
            \textbf{MAE$\downarrow$} & \textbf{MRE$\downarrow$} & \textbf{$\delta_1$$\uparrow$} &
            \textbf{Acc.$\uparrow$} & \textbf{mIoU$\uparrow$} & \textbf{Acc.$\uparrow$} &  \textbf{mIoU$\uparrow$} \\
            
    \midrule
    \midrule
        PanoFormer\hfilll~\cite{panoformer} & 14.5M & 11.8G     & .174     & .078      & 92.5      & .154       & .051      & 94.8  & 83.1       & 60.6   & 94.9      & 49.7   \\
        EGFormer\hfilll~\cite{egformer}     & 15.2M & 15.6G     & .170     & .075      & 93.1      & .150       & .049      & 95.2   & 86.5       & 66.4   & 95.0      & 51.5   \\
        SFSS\hfilll~\cite{panocmx}          & 15.1M & 18.9G     & .179     & .081      & 92.2      & .155       & .051      & 95.0  & 86.9       & 68.2  & 95.2      & 51.9  \\
    \midrule
        HexRUnet\hfilll~\cite{zhang2019orientation} & 14.0M & 12.4G     & .201 & .090        & 90.1 & -      & -     & -     & 81.7  & 56.1 & -      & - \\
        HealSWIN\hfilll~\cite{carlsson2024heal}     & 12.0M & 39.0G     & .189 & .084        & 92.2 & -      & -     & -     &  85.5 & 63.2 & -      & - \\
        Elite360D\hfilll~\cite{ai2024elite360d}     & 14.7M & 13.6G     & .169 & \best{.069} & 93.5 & .147  & .046  & 95.9  & 87.4  & 71.4 & 95.3   & 52.0 \\
    \midrule
        \textbf{\textit{OURS }} (\cref{tab:paramsours} (1)) & 14.9M  & 13.1G   & \best{.165}  & .071   & \best{94.0}   & \best{.142}  & \best{.045}     & \best{96.4}   & \best{88.6}    & \best{72.2}  & \best{95.8}      & \best{53.0}  \\
        \bottomrule
    \end{tabular}
    \caption{Quantitative comparison on sphere rank 7 (256$\times$512) resolution.}
  \label{tab:results_512}
\end{table*}

\subsection{Depth Estimation}\label{sec:depth}

\begin{figure*}[t]
\centering
\begin{tabular}{@{}c@{}c@{}c@{}c@{}c@{}c@{}c@{}c@{}}
    \includegraphics[width=0.162\linewidth, trim={0 660 1920 350}, clip]{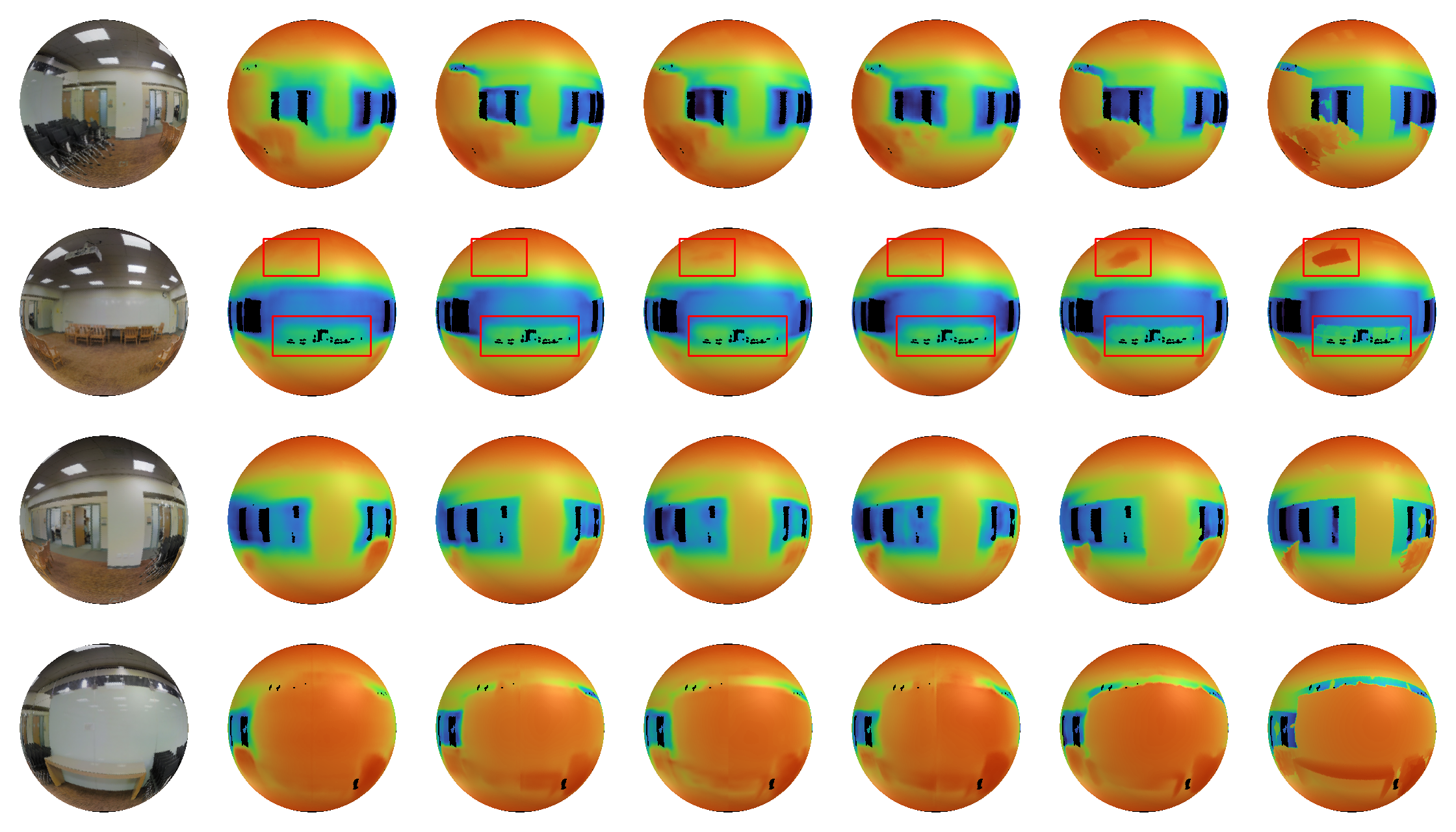} & 
    \includegraphics[width=0.162\linewidth, trim={320 660 1600 350}, clip]{images/very_new/stanford2d_depth.png} & 
    \includegraphics[width=0.162\linewidth, trim={640 660 1280 350}, clip]{images/very_new/stanford2d_depth.png} & 
    \includegraphics[width=0.162\linewidth, trim={1280 660 640 350}, clip]{images/very_new/stanford2d_depth.png} & 
    \includegraphics[width=0.162\linewidth, trim={1600 660 320 350}, clip]{images/very_new/stanford2d_depth.png} & 
    \includegraphics[width=0.162\linewidth, trim={1920 660 0 350}, clip]{images/very_new/stanford2d_depth.png} \\
    \includegraphics[width=0.162\linewidth, trim={0 30 1920 980}, clip]{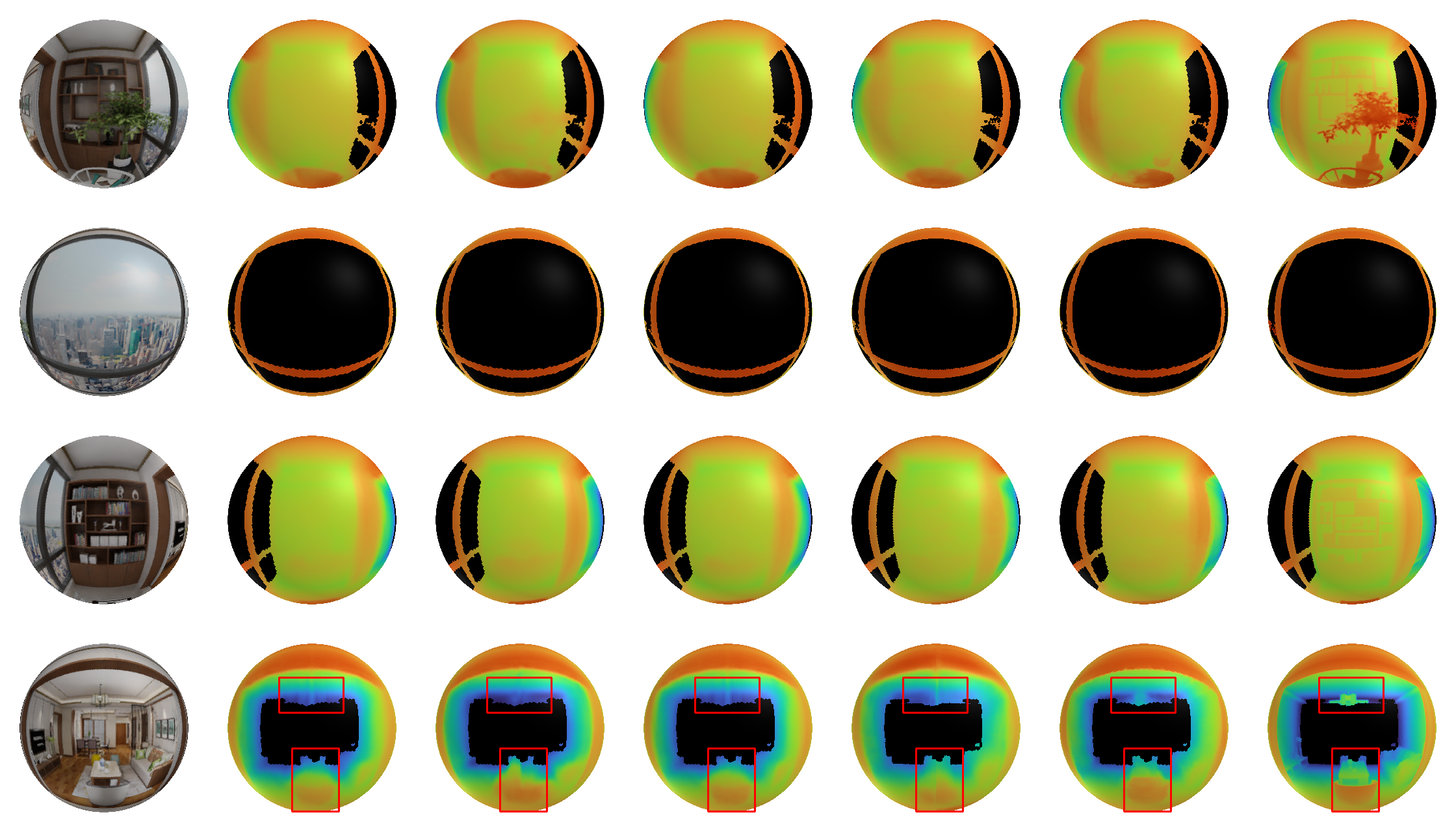} & 
    \includegraphics[width=0.162\linewidth, trim={320 30 1600 980}, clip]{images/very_new/structured_depth.png} & 
    \includegraphics[width=0.162\linewidth, trim={640 30 1280 980}, clip]{images/very_new/structured_depth.png} & 
    \includegraphics[width=0.162\linewidth, trim={1280 30 640 980}, clip]{images/very_new/structured_depth.png} & 
    \includegraphics[width=0.162\linewidth, trim={1600 30 320 980}, clip]{images/very_new/structured_depth.png} & 
    \includegraphics[width=0.162\linewidth, trim={1920 30 0 980}, clip]{images/very_new/structured_depth.png} \\
    RGB & PanoFormer & EGFormer & Elite360D & Ours & GT\\
\end{tabular}
\caption{{\bf Depth Estimation.} Top: Stanford2D3D. Bottom: Structured3D.}
\label{fig:depth}
\end{figure*}

In depth estimation, the task is a dense prediction of the pixels depths. We evaluated the depth prediction up to 10 meters for Stanford2D3D and 5 meters for Structured3D.
Following PanoFormer~\cite{panoformer}, we used the BerhuLoss~\cite{laina2016deeper} as a differentiable criterion to train the models. We then measured the Mean Absolute Error (MAE), Mean Relative Error (MRE), and $\delta_1$ accuracy, which are standard evaluation metrics for depth estimation.
Quantitative results on 256$\times$512 are reported in \cref{tab:results_512}. Our method is shown to outperform the baselines on all but one metric, where it was second. In evaluation on Stanford2D3D in  512$\times$1024 (\cref{tab:results_1024}), the gap is even larger, and our method significantly improved while the baselines did not.
\cref{fig:depth} presents sample results. Evidently, our results are considerably sharper than the baselines, especially around the center, and handle distortions around the poles better. We argue that this is due to spherical representation having a better effective resolution at the center, and does not suffer from distortions.
We have also found depth predictions of most baselines to have a boundary misalignment effect where the $360\degree$ and $0\degree$ meet, which our method has no such effect. This is further explored in the supplement.

\subsection{Semantic Segmentation}\label{sec:semantic}

\begin{figure*}[t]
\centering
\begin{tabular}{@{}c@{}c@{}c@{}c@{}c@{}c@{}c@{}c@{}}
    \includegraphics[width=0.162\linewidth, trim={0 660 1920 350}, clip]{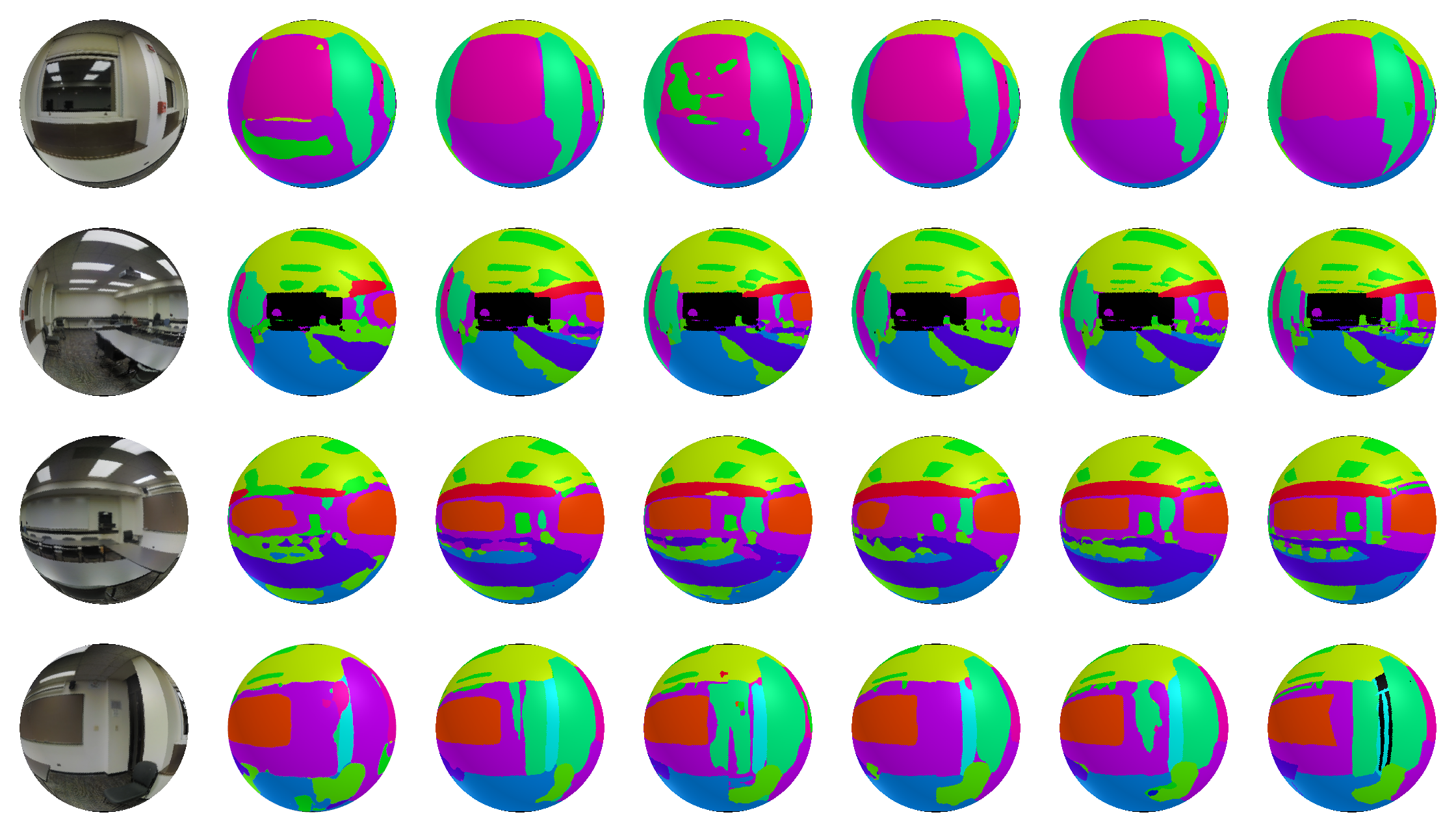} & 
    \includegraphics[width=0.162\linewidth, trim={320 660 1600 350}, clip]{images/very_new/stanford2d_semantic.png} & 
    \includegraphics[width=0.162\linewidth, trim={640 660 1280 350}, clip]{images/very_new/stanford2d_semantic.png} & 
    \includegraphics[width=0.162\linewidth, trim={1280 660 640 350}, clip]{images/very_new/stanford2d_semantic.png} & 
    \includegraphics[width=0.162\linewidth, trim={1600 660 320 350}, clip]{images/very_new/stanford2d_semantic.png} & 
    \includegraphics[width=0.162\linewidth, trim={1920 660 0 340}, clip]{images/very_new/stanford2d_semantic.png} \\
    \includegraphics[width=0.162\linewidth, trim={0 30 1920 980}, clip]{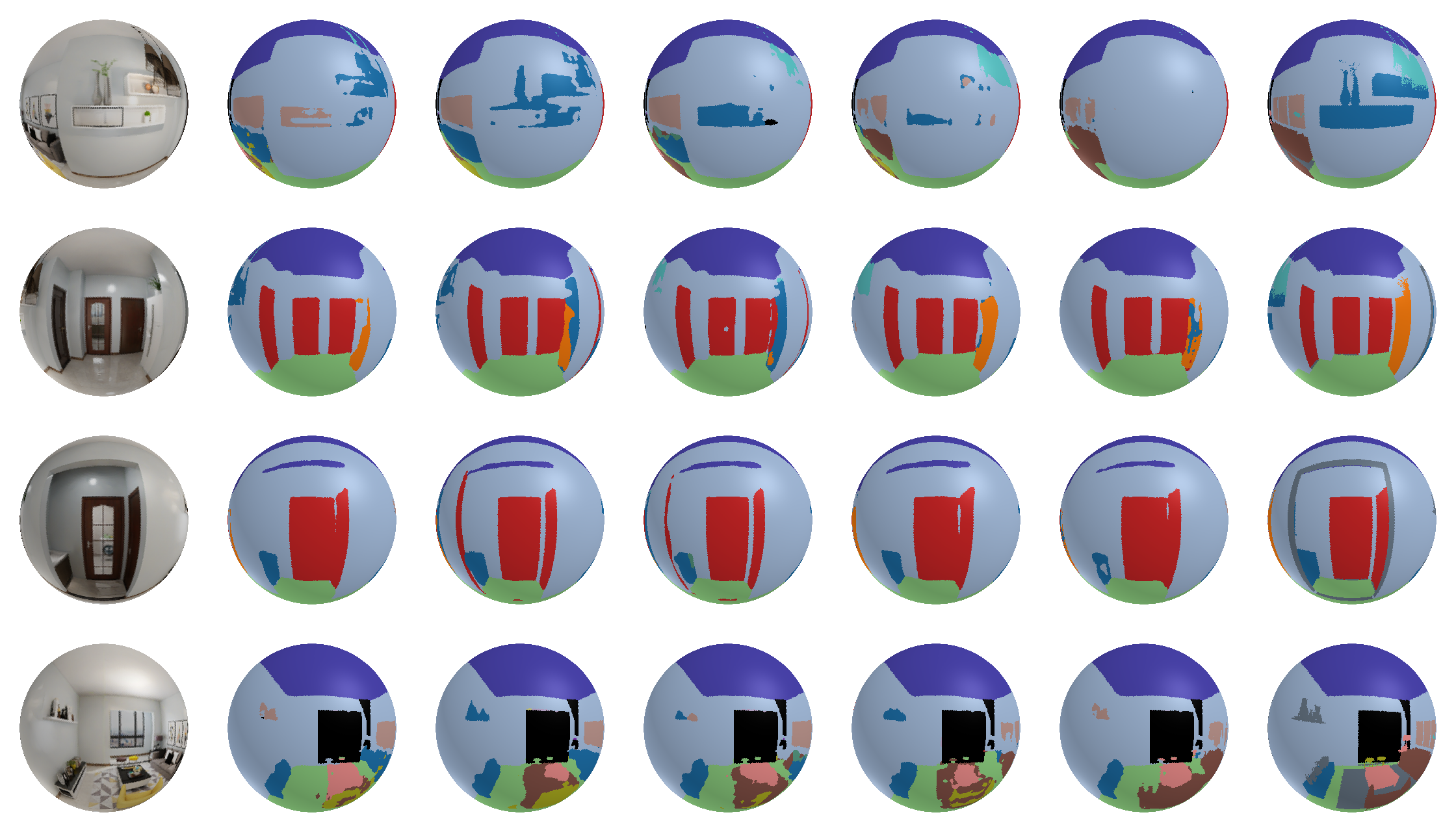} & 
    \includegraphics[width=0.162\linewidth, trim={320 30 1600 980}, clip]{images/very_new/structured_semantic.png} & 
    \includegraphics[width=0.162\linewidth, trim={640 30 1280 980}, clip]{images/very_new/structured_semantic.png} & 
    \includegraphics[width=0.162\linewidth, trim={1280 30 640 980}, clip]{images/very_new/structured_semantic.png} & 
    \includegraphics[width=0.162\linewidth, trim={1600 30 320 980}, clip]{images/very_new/structured_semantic.png} & 
    \includegraphics[width=0.162\linewidth, trim={1920 30 0 980}, clip]{images/very_new/structured_semantic.png} \\
    RGB & PanoFormer & EGFormer & Elite360D & Ours & GT\\
\end{tabular}
\caption{{\bf Semantic Segmentation.} Top: Stanford2D3D. Bottom: Structured3D.}
\label{fig:semantic}
\end{figure*}

In semantic segmentation, the task is to predict a semantic class per image pixel. Stanford2D3D has 13 categories, while Structured3D has 40. We used the standard Categorical Cross Entropy loss, where the background category was ignored. For evaluation, we measured the global pixel accuracy and the mIoU percentage, which are standard metrics for semantic segmentation.
Results are depicted in \cref{tab:results_512}. In this task as well, our method outperformed the baselines on all metrics. \cref{fig:semantic} presents typical samples. Similarly, our semantic segmentation results present the same improvement in terms of distortion handling and improvement around the center of the image we have observed for depth estimation. 

\begin{table}[t]
    \centering
    \begin{tabular}{@{}l@{~}|@{~}c@{~~}c@{~}|@{~}c@{}c@{}c@{}}
    \toprule
    \textbf{Model} & \textbf{Params} & \textbf{Flops} & \textbf{MAE$\downarrow$} & \textbf{MRE$\downarrow$} & \textbf{$\delta_1$$\uparrow$} \\
    \midrule
    \midrule
    PanoFormer  & 14.5M & 44.7G     & .167 & .072 & 93.7\\
    Elite360D   & 14.7M & 51.4G     & .181 & .077 & 93.2\\
    \midrule
    \textbf{\textit{OURS }}~(\cref{tab:paramsours} (2)) & 14.9M & 52.7G & \best{.147} & \best{.065} & \best{94.0} \\
    \bottomrule
    \end{tabular}
    \caption{Depth estimation on Stanford2D3D. rank 8 (512$\times$1024).}
  \label{tab:results_1024}
\end{table}

\subsection{Ablation Study}\label{sec:ablation}

To explore the contribution of each one of the elements we have conducted several ablation studies in which multiple variants of our method are tested. These tests were conducted in the context of depth estimation on Stanford2D3D.

\paragraph{Positional Encoding}

Two main components of the method are the global absolute and relative bias position encodings. The vertical global position encoding injects vertical position while remaining rotation equivariant horizontally. The relative bias within the self-attention provides relative information. We compared various configurations that omit one or both of the encodings. Results can be seen in \cref{tab:ablation_position}. Evidently, the absence of positional encoding, either global or relative, drastically impacts the performance of the model. Also, relative positional bias proved more critical than the global one.

\paragraph{Scaling Coefficients}

Two key modifiable scaling coefficients, the ``window'' ($C_{win}$) and ``head'' ($C_{head}$), affect the capacity of the attention module. The former affects the size of the attention window to incorporate further nodes in the graph. The latter increases the capacity of the attention head by increasing the head dimension.
We tested the effect of these coefficients by gradually increasing them separately and together.
As shown in \cref{tab:ablation_window}. Increasing the window up until $C_{win}=2$ improves the performance. Interestingly, further increase of the window did not show an improvement. This could suggest that such large windows interfere with convergence. As expected, increasing the attention head also improves results, even up to $C_{head}=4$. Unsurprisingly, this suggests that the model could benefit from an overall increase in parameters, which is beyond the scope of this work.

\begin{table}
    \centering
    \begin{tabular}{@{}lcc|cc@{}}
    \toprule
    Variant & AbsPos & RelPos & \textbf{MAE$\downarrow$} & \textbf{MRE$\downarrow$} \\
    \midrule
    \midrule
      \textit{No Pos. Enc.}         & \xmark      & \xmark      & .326     & .094  \\ 
      \textit{No Rel. Pos. Enc.}    & \cmark      & \xmark      & .251     & .091  \\ 
      \textit{No Abs. Pos. Enc.}    & \xmark      & \cmark      & .218     & .088  \\ 
      \textit{With Pos. Enc.}       & \cmark      & \cmark      & \bf.189     & \bf.077  \\ 
        \bottomrule
    \end{tabular}
    \caption{Ablation Study: Positional Encoding.}
    \label{tab:ablation_position}
~\\
    \centering
    \begin{tabular}{@{}l@{~~~~~~~~~}c@{~~~~~~}c@{~~~~}|cc@{}}
    \toprule
    Variant & $C_{head}$ & $C_{win}$ & \textbf{MAE$\downarrow$} & \textbf{MRE$\downarrow$} \\
    \midrule
    \midrule
    \textit{Base variant}       & 1     & 1     & .189      & .077  \\ 
    \textit{No window}          & 1     & 0     & .412      & .122  \\ 
    \textit{Medium window}      & 1     & 2     & .175      & .073 \\ 
    \textit{Large window}       & 1     & 3     & .180      & .078 \\ 
    \textit{Medium head}        & 2     & 1     & .171      & .075  \\
    \textit{Large head}         & 4     & 1     & \bf.162      & \bf.067  \\ 
    \textbf{\textit{Conf. (1) \cref{tab:paramsours}}}   & 2        & 2        & .165  & .071 \\ 
        \bottomrule
    \end{tabular}
    \caption{Ablation Study: Head and Window Coefficients.}
    \label{tab:ablation_window}
\end{table}

\section{Discussion and Limitations} \label{sec:discussion}

This paper introduces a pioneering approach to omnidirectional 360$\degree$ perception, leveraging a geodesic polyhedron and a unique architecture designed for spherical computation. However, there is ample room for optimization and enhancement of its features. The potential for improvement spans various aspects of the model, from its core architecture to its computational efficiency.

One of the promising avenues for future work lies in exploring additional domains where this architecture could be applied. Beyond the realms of depth estimation and semantic segmentation in 360$\degree$ imagery, spherical data representation is also prominent in other areas such as medical imaging, where spherical structures are common, and geographical data analysis, which often deals with global-scale data that naturally fit a spherical model. The adaptability of our approach to these varied domains could pave the way for groundbreaking applications and methodologies.

While the current work has focused on supervised classification and regression tasks, extending this model to generative models presents an exciting frontier. Generative models could leverage the spherical data representation for creating highly realistic and comprehensive 3D environments among other applications. The unique challenges of generative tasks on spherical domains, such as ensuring continuity and avoiding distortions, call for accurate modeling of the type we present here.

Another area for improvement is the computational efficiency of the model, particularly regarding its GPU usage. Despite the number of parameters and flops being on par with baseline models, our approach exhibits slower performance (\tildemid30\% slower than PanoFormer). This is largely because the gathering operations on the sphere graph are not optimized with dedicated CUDA functions in PyTorch, leading to less efficient execution. Addressing this engineering challenge could significantly enhance the model's appeal by reducing both the GPU footprint and runtime. 

\section{Conclusion} \label{sec:conclusion}

We have presented a novel method for omnidirectional perception that successfully utilizes spherically represented data. 
The method addresses the underlying spherical structure, which comes up in three different places: in the down and upsampling operators, in the positional encoding, and, above all, in the definition geometric representation and that of the neighborhoods for the local attention operator.

Our proposed method relies on the icosphere representation and utilizes its favorable properties to derive sphere-based operators. Using these, we surpass the state of the art in omnidirectional perception in both depth estimation and semantic segmentation.

\newpage

{
    \small
    \bibliographystyle{ieeenat_fullname}
    \bibliography{main}
}

\clearpage
\setcounter{page}{1}

\maketitlesupplementary
\appendix
\makeatletter
\begin{bibunit}

\section{Technical Details}\label{supp:tech}

In this section, we describe the implementation of our method in detail.

\subsection{Architecture}
Given a group of nodes $G_r$ of rank $r$ on a sphere representation, an input data of rank $r_0$ is projected by an ``Input Projection'' layer, which is a fully connected layer mapping each node value to $D_1=32$ dimensions, then followed by GELU activation. In addition, an initial center downsampling layer reduces the rank of the data to $r_1=r_0-1$. Global position encoding, which is explained in \cref{supp:tech/abs_pos}, of dimension $D_1$ is added, forming the final embedded data.
\begin{gather}
e^0_i = \text{GELU}(\text{Proj}_{in}(x_i)), \forall i \in G_{r_0} \\
\bar{e}^1 = \text{POOL}(e^0) \\
e^1_i = \bar{e}^1_i + \text{Pos}_0(\phi_i),  \forall i \in G_{r_1}
\end{gather}

The encoder consists of $N=4$ consecutive spherical attention modules (SAM). Each consists of $M=2$ spherical attention blocks (SAB). For each $n \in [1 .. 4]$, the embedding dimension is $D_n = 2^{i-1} \cdot D_1$, resulting in 32, 64, 128, 256.
At the end of each SAM is a downsampling block, which consists of a center downsampling layer, followed by a fully connected layer that projects the embedding from $D_n$ to $D_{n+1}$. The downsampling reduces the sphere dimension from $G_{r_0-n}$ to $G_{r_0-n-1}$.
Therefore, for all $m,n \in [1..M]\times[1..N]$:
\begin{gather}
e^{n,m} = \text{SAB}_{n,m}(e^{n,m-1})\label{eq:sab} \\
e^{n+1} = \text{Proj}_n(\text{POOL}(e^{n,M}))
\end{gather}
, where $e^{n,0}$ is $e^n$ produced by the previous module.

As shown in \cref{fig:slsa} (in the main paper), each SAB is composed of a spherical local self attention (SLSA) with a residual connection, and an MLP that consists of 2 fully connected layers with in/out dimensions $D_n$ and hidden dimension $4 \cdot D_n$, and a GELU activation.
\begin{gather}
    \bar{h}^{n,m} = h^{n,m-1} + \text{SLSA}_{n,m}(h^{n,m-1}) \\
    h^{n,m} = \bar{h}^{n,m} + \text{MLP}_{n,m}(\text{LayerNorm}(\bar{h}^{n,m}))
\end{gather}

As can be seen in \cref{fig:arch}. The bottleneck follows the same structure as \cref{eq:sab}, operating on $G_{r_0-N-1}$ with $D_{N+1}=512$, and without a downsampling layer.
The decoder follows the same structure as the encoder, with the slight modification that an upsampling with projection to $D_n$ is performed instead of downsampling, and that the encodings are merged with a bypass connection from the encoder with the same sphere rank.
For the decoder:
\begin{gather}
    d^{n,0} = \text{UNPOOL}(\text{Proj}_n(d^{n+1})) \\
    d^{n,0} = \text{CONCAT}(d^{n,0} , e^{n,M}) \\
    d^{n,m} = \text{SAB}_{n,m}( d^{n,m-1}) 
\end{gather}
, where $d^{n} = d^{n,M}$. The concatenation also means that a decoder stage has a dimension of $2D_n$. Which is twice the size of the encoder.

Finally, a fully connected ``output projection'' projects the encoding $d^{1}$ from 64 channels to 1 for depth, and the number of classes in segmentation.
\begin{gather}
    y = \text{Proj}_{out}(\text{UNPOOL}(d^{1}))
\end{gather}

\subsection{Spherical Local Self Attention}\label{supp:tech/slsa}
The spherical local self attention (SLSA)~\cref{fig:slsa} (in the main paper) is an attention layer that adheres to the sphere structure. A local operation is performed between neighboring nodes. A hyperparameter $C_{win}$ controls how far through the graph neighbors are considered.

An input $e$ with dimension $D$ is projected into \{$q$,$k$,$v$\} vectors. All have the same size, equal to $C_{head}\cdot D$. They are then split into $H$ heads. Per stage, the encoder has 2,4,8,16 heads, the decoder has 4,8,16,16, and the bottleneck 16. Absolute position encoding is again added to $q$ and $k$, then the vectors are normalized.
Based on the graph structure $G$, for each $e_i$, we denote $E^K_i$ as the set of neighbors given $C_{win}=K$.
Then, for each node $i \in G$, the dot product between the query $q_i$ and each key in its neighborhood. Relative position bias is added to each dot product result. A final softmax operation normalizes the attention weights. This is applied per attention head, but for conciseness, the head notation is not added.
\begin{gather}
    \bar{a}_{i,j} = q_i \cdot k_j + \text{RelPos}(i,j), \forall i \in G, \forall j \in E^K_i \\
    \{a_{i,1},..a_{i,|E^K_i|}\} = \text{Softmax}(\{\bar{a}_{i,1},..\bar{a}_{i,|E^K_i|}\})
\end{gather}

The attention map is then multiplied by the value vectors $v_j \forall j \in E^K_i$ to form the output of the head.
\begin{gather}
    \bar{o}_{i} = a_{i,j} \cdot v_j, \forall i \in G, \forall j \in E^K_i
\end{gather}

Finally, the different heads are concatenated and a linear output projection projects the concatenated heads to the input dimension.
\begin{gather}
    o_{i} = \text{Proj}(\text{CONCAT}(\{\bar{o}^h_i\}^H_{h=1}))
\end{gather}
The encoder and decoder use 2,4,8,16 heads per equivalent stage $n \in [1..N]$.

\subsection{Global Positional Encoding}\label{supp:tech/abs_pos}
Our method employs a vertical global positional encoding to inform the model of each node's vertical position on the sphere. The vertical position is provided by the angle $\phi\in[0, \pi]$, where 0 denotes the top point with $z=1$ and $\pi$ denotes the bottom point with $z=-1$.

To encode the position. The angle value $\phi$ is encoded with $D$ sine and cosine wave functions of different frequencies, resulting in a total of $2D$ position features. The frequencies are sampled between $F_{min}=1$ (fixed) and $F_{max}$ (configurable) by the function:
\begin{equation}
    f_i = F_{max} ^ {i/(D-1)},
\end{equation}
where i is a value between 0 and $D-1$.

The position vector is then defined as:
\begin{equation}
    \begin{split}
    \hat{\phi} &= (2\phi-\pi) \\
    pos(\phi) &= [\sin(f_i \hat{\phi})|_{i=0}^{D-1} , \cos(f_i \hat{\phi})|_{i=0}^{D-1}]
    \end{split}
\end{equation}

\begin{figure}[h]
    \centering
    \includegraphics[width=\linewidth, trim={40 90 40 100}, clip]{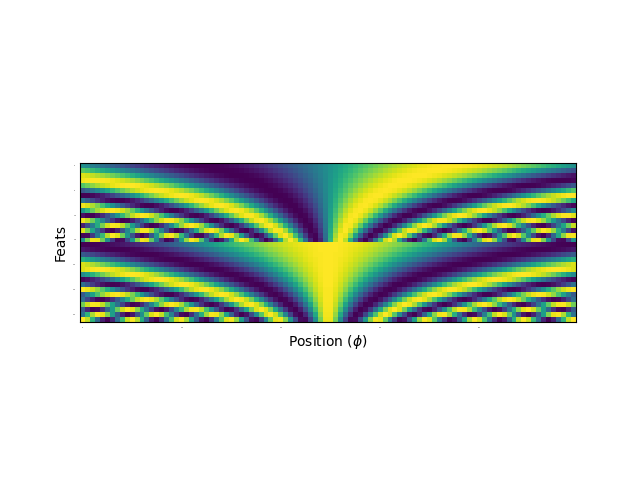}
    \caption{Visualizing the global position encoding per angle $\phi$ from $0$ to $\pi$.}
    \label{supp:fig:posenc}
\end{figure}

We used $D=16$, $F_{max}=10$. Fig.~\ref{supp:fig:posenc} illustrates the global position encoding.

As we apply global position encoding in our model in different stages, we take the sinusoidal position encodings and project them with a learned linear layer to the dimension of the target vector, which can vary based on the stage within the architecture.

\subsection{Relative Position Encoding}\label{supp:tech/rel_pos}
The relative position encoding has two steps. First, the relative angles $(\Delta\theta, \Delta\phi)$ are computed for each pair of neighboring nodes. Second, using the relative angles, a learned value is sampled from a 7$\times$7 grid.

For a graph representation $G$, with K-order neighbor mapping $E^K_i$ for each node $i \in G$. The angles $(\Delta\theta, \Delta\phi)$ are computed between each node $i$ and each node $j \in E^K_i$. This is done by rotating the nodes based on $\theta_i, \phi_i$, and extracting the new $\Delta\theta_j, \Delta\phi_j$ after rotation. The rotated angles are computed by converting the nodes' positions to cartesian form $(x,y,z)$, then multiplying by a rotation matrix, and converting back to spherical form.
\begin{gather}
    x_j, y_j, z_j = \text{ToCartesian}(\theta_j, \phi_j)\\
    [\hat{x}_j, \hat{y}_j, \hat{z}_j]^T = R_i*[x_j, y_j, z_j]^T \\
    \Delta\theta_j, \Delta\phi_j  = \text{ToSpherical}(\hat{x}_j, \hat{y}_j, \hat{z}_j)
\end{gather}
The rotation matrix is defined by $\phi_i, \theta_i$ as follows:
\begin{gather}
    R_i = R^\phi_i * R^\theta_i\\
    \hat{\theta}_i = \theta - \pi, \hat{\phi}_i = \phi - \frac{\pi}{2}\\
    R^\phi_i = \begin{bmatrix}
                \cos(-\hat\phi_i) & 0 & \sin(-\hat\phi_i) \\
                0 & 1 & 0 \\
                -\sin(-\hat\phi_i) & 0 & \cos(-\hat\phi_i)
                \end{bmatrix}\\
    R^\theta_i = \begin{bmatrix}
                \cos(-\hat\theta_i) & -\sin(-\hat\theta_i) & 0 \\
                \sin(-\hat\theta_i) & \cos(-\hat\theta_i) & 0 \\
                0 & 0 & 1
                \end{bmatrix}
\end{gather}
After the rotation, node $i$ receives $\Delta\theta_i=0, \Delta\phi_i=0$, while the neighbors on the right and left receive positive and negative (respectively) $\Delta\theta_j$ and above and below receive negative and positive (respectively) $\Delta\phi_j$.
This computation is performed only once in advance, as the structure of the graph is fixed.
After rotation, we use bilinear interpolation to sample the value in a 7$\times$7 grid according to $\Delta\theta_j, \Delta\phi_j$. As was shown in \cref{fig:rel_pos} (in the main paper). Since the grid is in the range [-1,1] on both axes, we normalize $\Delta\theta_j, \Delta\phi_j$ by the max absolute value of each angle coordinate over all $i \in G$. The value is added as bias to the attention map.

\section{Training Protocol}\label{supp:protocol}
We trained each model for a similar amount of iterations. For Stanford2D3D~\cite{armeni2017joint} we used 25K iteration. For Structured3D~\cite{structured3d}, which is a much larger dataset, we trained for 160K iterations. Each iteration has a batch size 16.
The models were trained with the Adam~\cite{kingma2014adam} optimizer, with a learning rate of 1e-4.
Evaluation on the validation set was performed every 400 iterations, monitoring the best score. 

\subsection{Fast Resolution Upscaling}\label{supp:protocol/high_res}
\begin{figure}
    \centering
    \includegraphics[width=\linewidth, trim={0, 0, 45, 20}, clip]{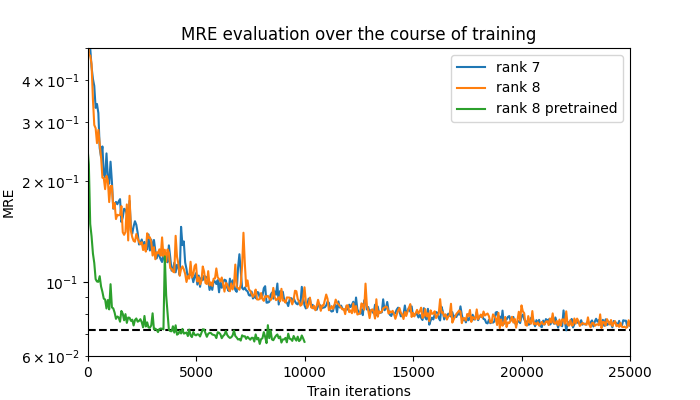}
    \caption{\textbf{High resolution training comparison.} Training a model in rank 7 and rank 8 resolutions converge at the same rate. However, fine-tuning a high resolution model on pretrained weights of a lower resolution converges much faster.}
    \label{supp:fig:train_high_res}
\end{figure}

Since the model is composed of local operations. A model can be finetuned for higher resolution with pretrained weights trained on a lower resolution. As long as the hyperparameters of the network have not changed.
We have found this to significantly speed up training, and convergence occurred in much fewer iterations. Since training on high resolution is much slower per iteration, this is a huge benefit. ~\cref{supp:fig:train_high_res} shows the MRE throughout training on Stanford2D3D, for a rank 7 model, rank 8 model, and rank 8 model pretrained on rank 7 weights.
As can be seen, when given enough iterations, rank 8 reaches rank 7's performance. But, a pretrained model on rank 7 converged much faster and already surpassed its counterparts in less than 5K iterations.

\section{Performance Analysis}\label{supp:performance}
In this section, we performed a deeper analysis of our method's performance.

\subsection{Boundary Effect}
\begin{figure}
    \centering
    \begin{tabular}{@{}c@{}c@{}c@{}c@{}}
    \includegraphics[width=.25\linewidth, trim={20 0 1940 960}, clip]{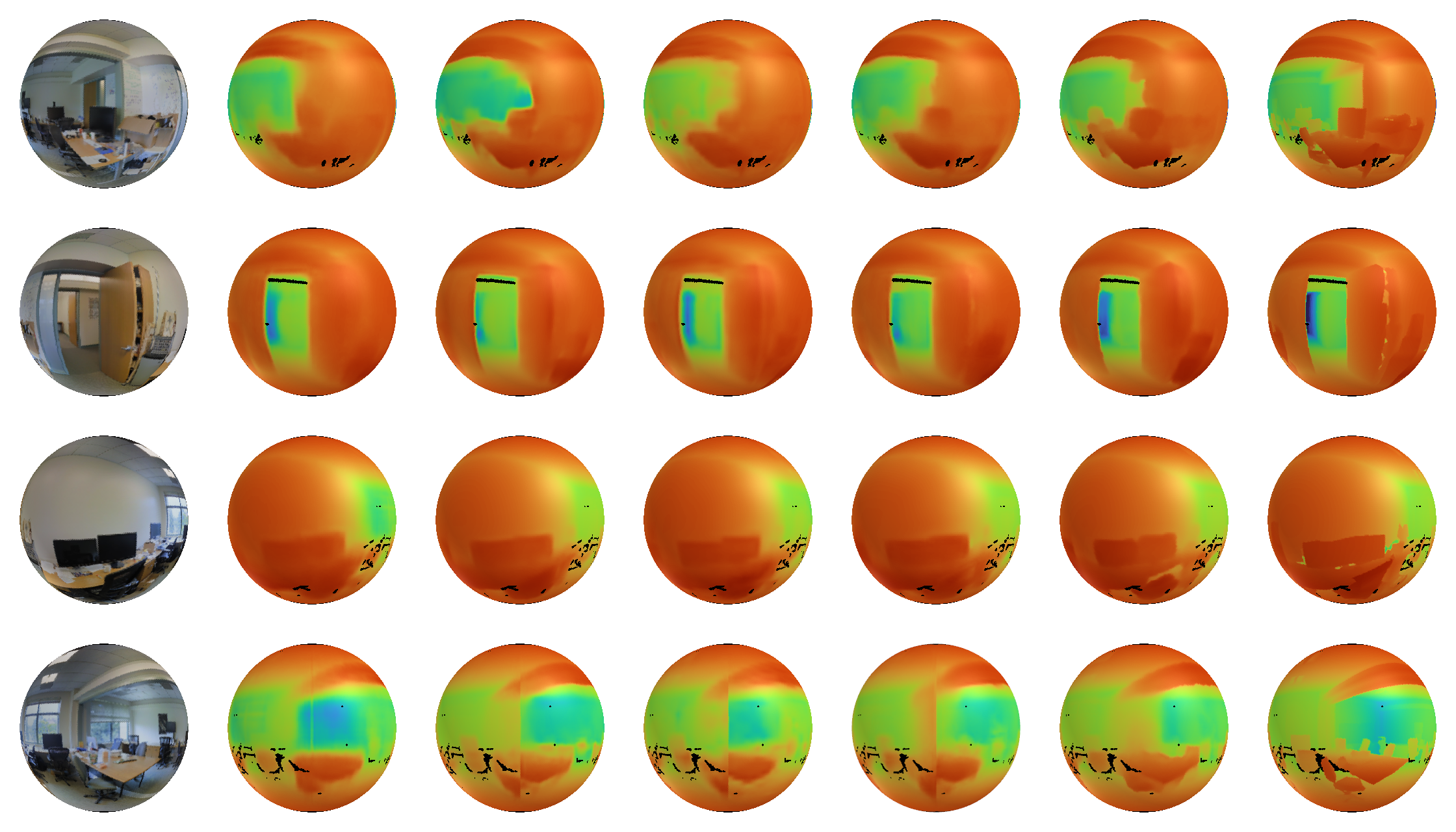} &
    \includegraphics[width=.25\linewidth, trim={340 0 1620 960}, clip]{images/supp/boundary1.png} &
    \includegraphics[width=.25\linewidth, trim={660 0 1300 960}, clip]{images/supp/boundary1.png} &
    \includegraphics[width=.25\linewidth, trim={980 0 980 960}, clip]{images/supp/boundary1.png} \\
    RGB & PanoFormer & EGFormer & SFSS \\
    &
    \includegraphics[width=.25\linewidth, trim={1300 0 660 960}, clip]{images/supp/boundary1.png} &
    \includegraphics[width=.25\linewidth, trim={1620 0 340 960}, clip]{images/supp/boundary1.png} &
    \includegraphics[width=.25\linewidth, trim={1940 0 20 960}, clip]{images/supp/boundary1.png} \\
    & Elite360D & OURS & GT  \\
    \includegraphics[width=.25\linewidth, trim={20 0 1940 960}, clip]{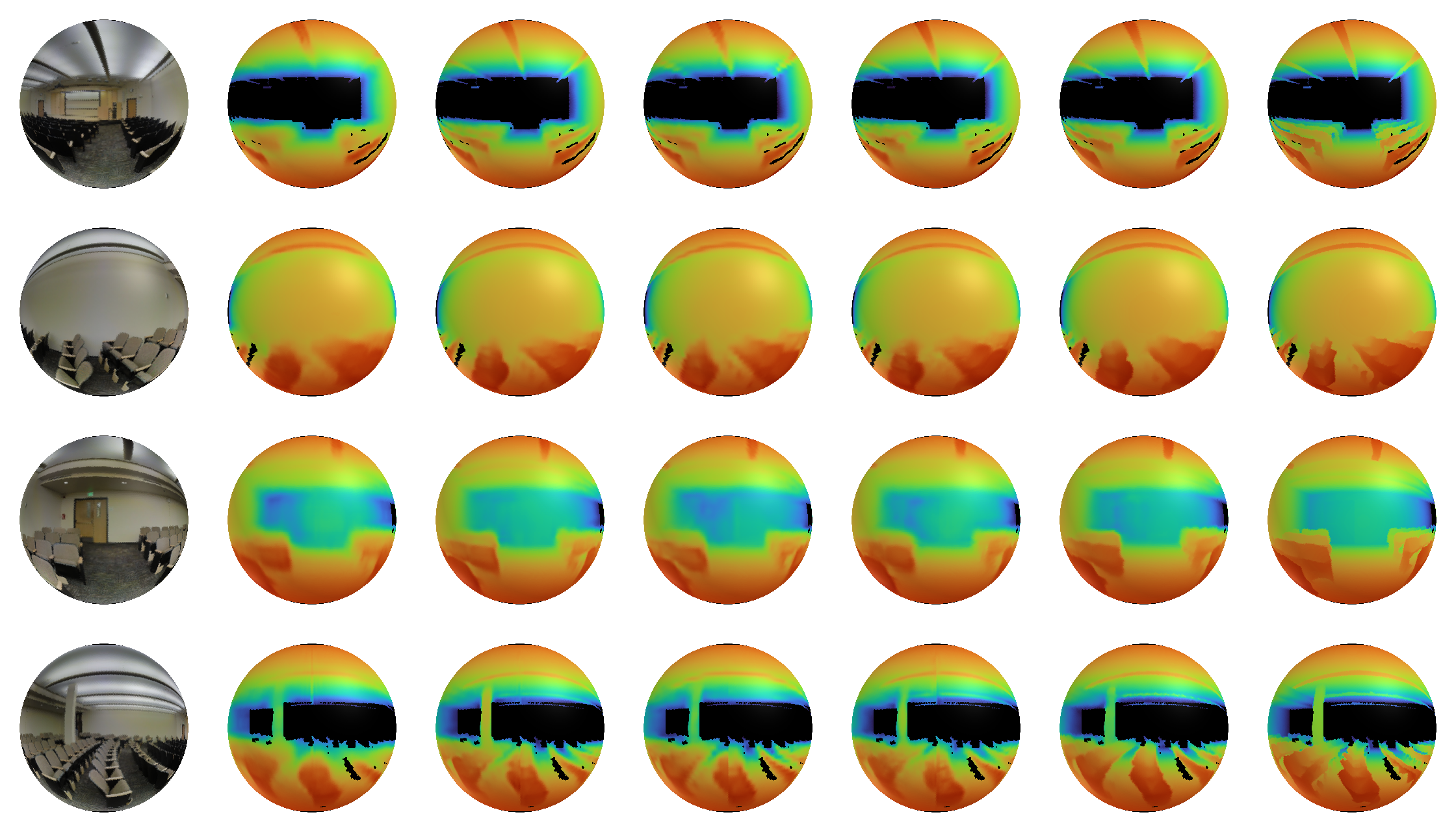} &
    \includegraphics[width=.25\linewidth, trim={340 0 1620 960}, clip]{images/supp/boundary2.png} &
    \includegraphics[width=.25\linewidth, trim={660 0 1300 960}, clip]{images/supp/boundary2.png} &
    \includegraphics[width=.25\linewidth, trim={980 0 980 960}, clip]{images/supp/boundary2.png} \\
    RGB & PanoFormer & EGFormer & SFSS \\
    &
    \includegraphics[width=.25\linewidth, trim={1300 0 660 960}, clip]{images/supp/boundary2.png} &
    \includegraphics[width=.25\linewidth, trim={1620 0 340 960}, clip]{images/supp/boundary2.png} &
    \includegraphics[width=.25\linewidth, trim={1940 0 20 960}, clip]{images/supp/boundary2.png} \\    
    & Elite360D & OURS & GT  \\
    \end{tabular}
    \caption{\textbf{Boundary effect comparison.} All baselines show a boundary issue at the 0$\degree$ horizontal angle.}
    \label{supp:fig:boundary}
\end{figure}

In our analysis of the baselines, we observed a recurring boundary effect at the points where the horizontal 360$\degree$ and 0$\degree$ meet. This boundary effect is most noticeable in depth estimation where the estimated depth is not continuous there. It can come as a misalignment between the two sides of the image, or as a vertical line that passes through the image. 
This boundary effect did not exist in our model, since it is fully horizontally equivariant. This is illustrated in \cref{supp:fig:boundary}, where we oriented the images to have the 0$\degree$ at the center. It can be seen that all baselines suffer from this boundary issue to some extent, while our method does not.

\subsection{Error Distribution on the Sphere}
\begin{figure}
    \centering
    \begin{tabular}{@{}c@{}c@{}c@{}c@{}}
    \includegraphics[width=0.25\linewidth, trim={40 50 1960 50}, clip]{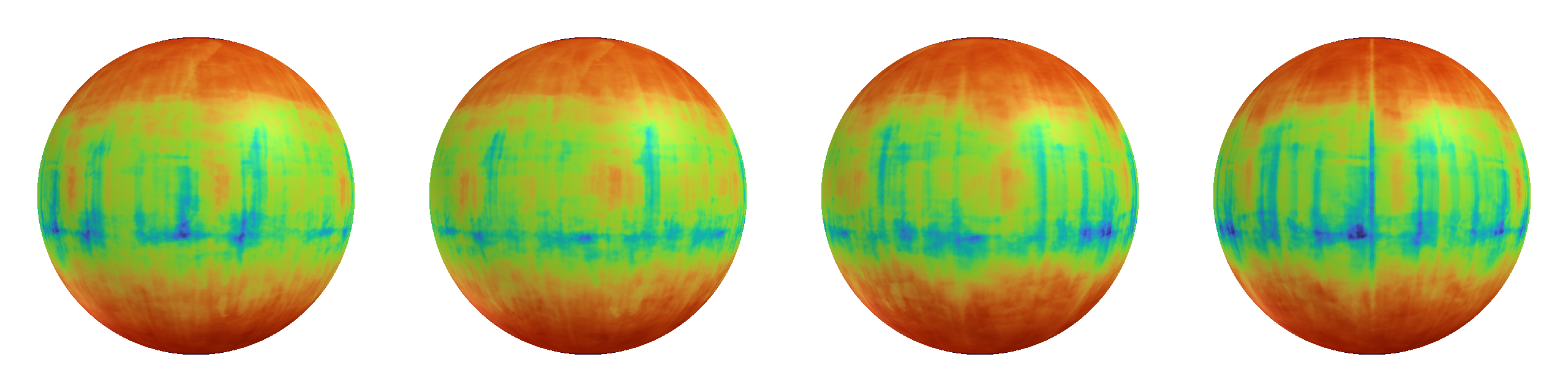} &
    \includegraphics[width=0.25\linewidth, trim={680 50 1320 50}, clip]{images/supp/error/pano.png} &
    \includegraphics[width=0.25\linewidth, trim={1320 50 680 50}, clip]{images/supp/error/pano.png} &
    \includegraphics[width=0.25\linewidth, trim={1960 50 40 50}, clip]{images/supp/error/pano.png} \\
    \includegraphics[width=0.25\linewidth, trim={40 50 1960 50}, clip]{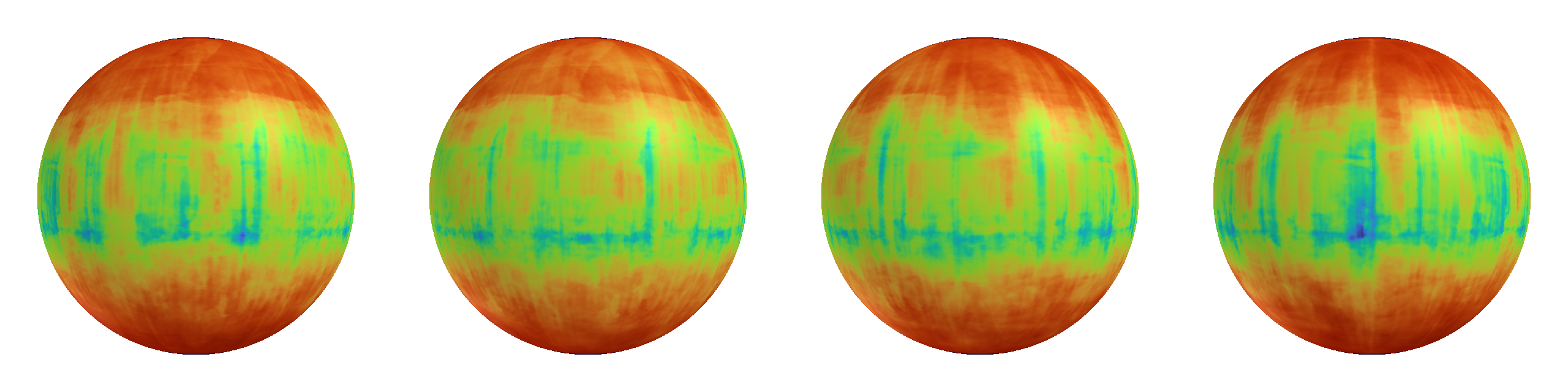} &
    \includegraphics[width=0.25\linewidth, trim={680 50 1320 50}, clip]{images/supp/error/egformer.png} &
    \includegraphics[width=0.25\linewidth, trim={1320 50 680 50}, clip]{images/supp/error/egformer.png} &
    \includegraphics[width=0.25\linewidth, trim={1960 50 40 50}, clip]{images/supp/error/egformer.png} \\
    \includegraphics[width=0.25\linewidth, trim={40 50 1960 50}, clip]{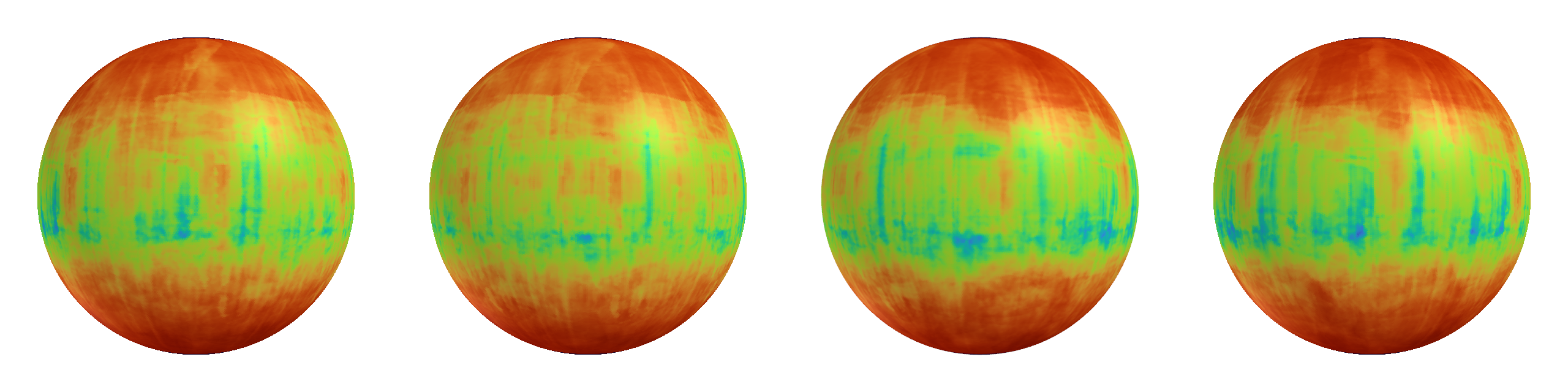} &
    \includegraphics[width=0.25\linewidth, trim={680 50 1320 50}, clip]{images/supp/error/sphere.png} &
    \includegraphics[width=0.25\linewidth, trim={1320 50 680 50}, clip]{images/supp/error/sphere.png} &
    \includegraphics[width=0.25\linewidth, trim={1960 50 40 50}, clip]{images/supp/error/sphere.png} \\
    \end{tabular}
    \caption{\textbf{MAE per location on the sphere.} \textbf{From top to bottom:} PanoFormer, EGFormer, Ours. \textbf{From left to right:} different sides of the 360$\degree$ view. Dark red values indicate low error, while bright blue values indicate high error.}
    \label{supp:fig:radial_error}
\end{figure}

To better understand the strengths and weaknesses of each method. We also analyzed the error per location on the sphere.
This is illustrated in \cref{supp:fig:radial_error}, where we visualize the average error of many depth predictions over the validation set, normalized by the max error of all models.
The visualization shows a low average error in red, and a high one in blue.
As can be seen, our method has less error at the bottom and top (noticed by the darker red tone), due to the better handling of the distortions, while also having less error at the center (fewer bright spots), due to its higher effective resolution at the center. The higher effective resolution is a result of distributing the data points uniformly on the sphere, instead of the unbalanced sampling of ERP in favor of the top and bottom of the image.

\begin{figure*}[t]
    \centering
    \includegraphics[width=\linewidth, trim={0 690 0 1330}, clip]{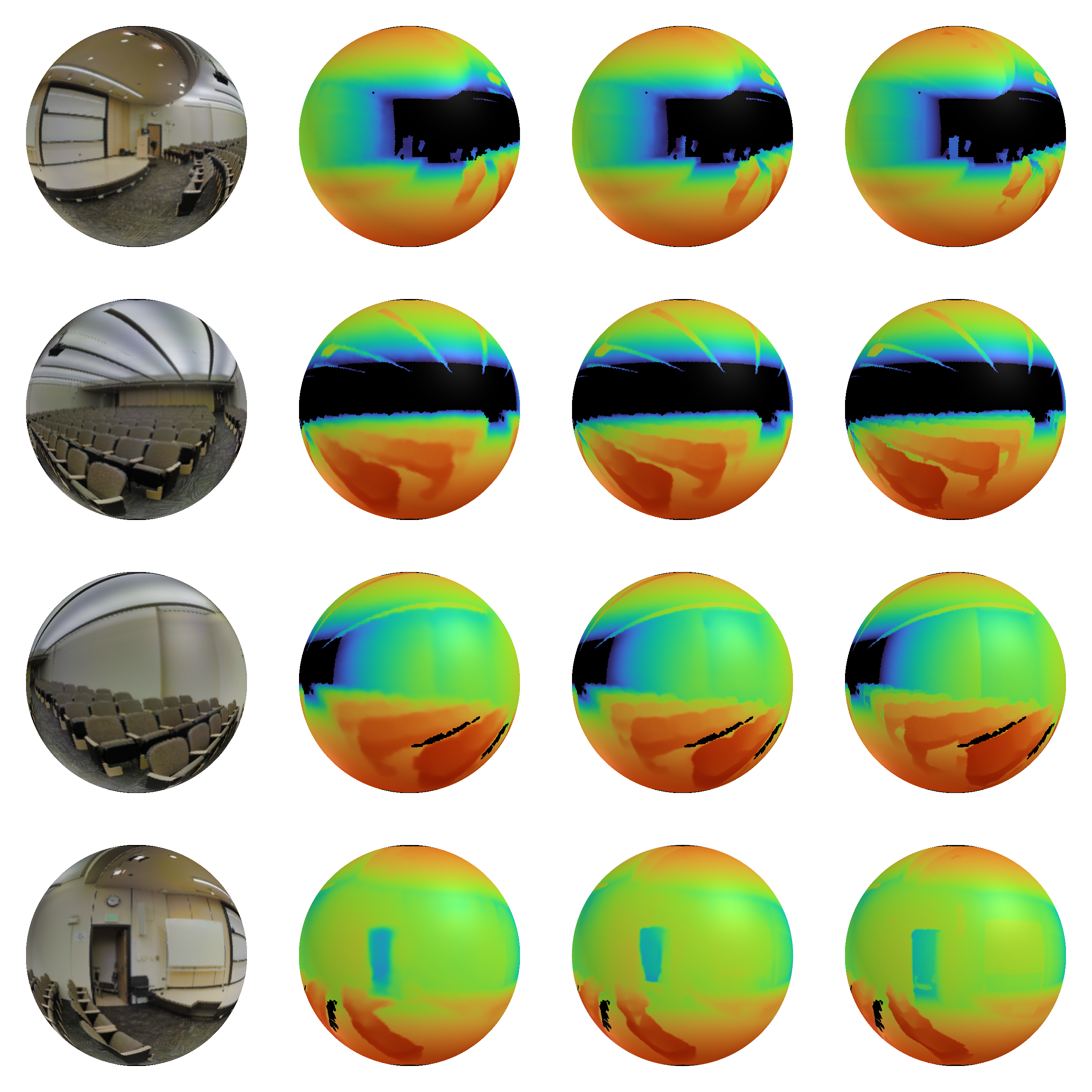} \\ 
    \includegraphics[width=\linewidth, trim={0 1330 0 690}, clip]{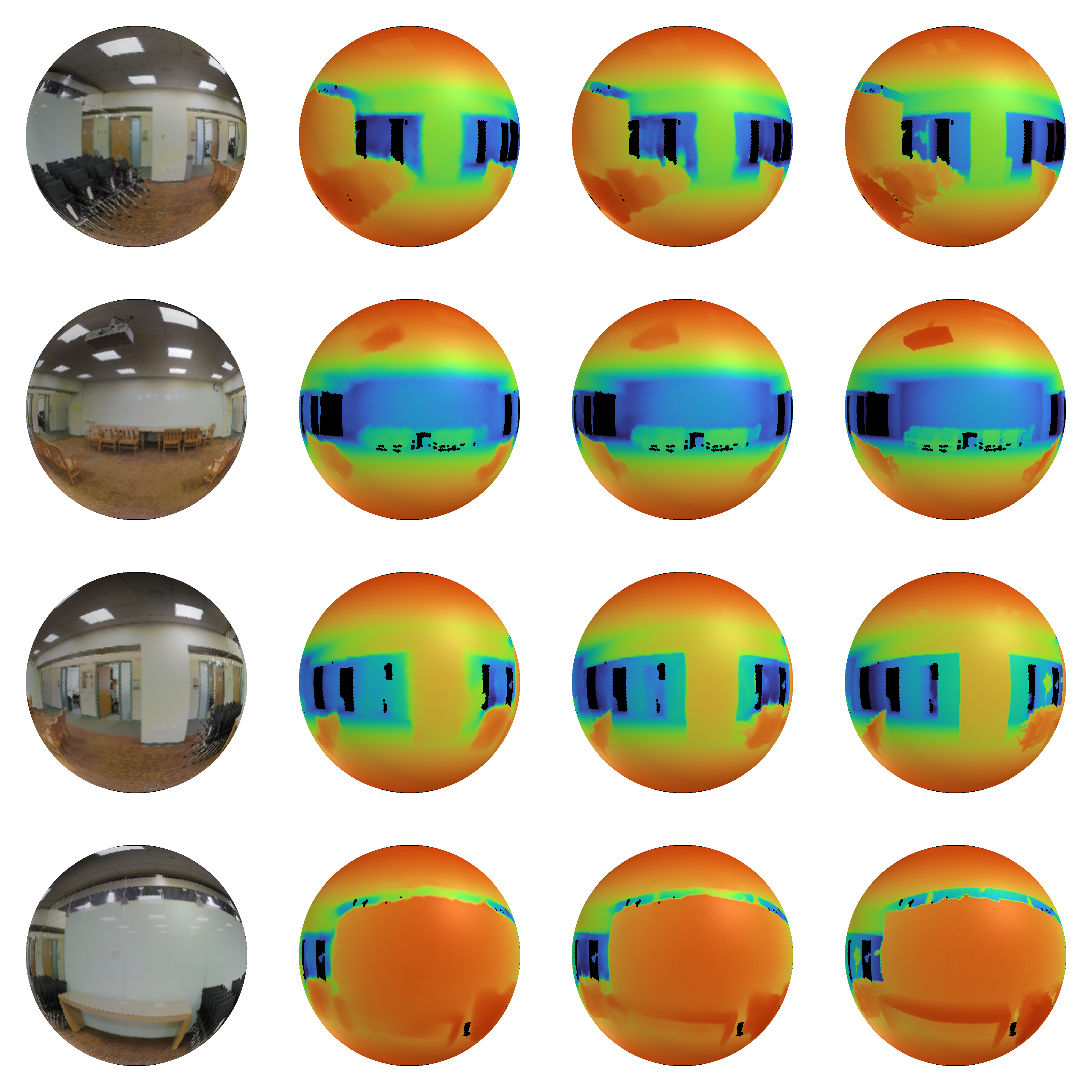} \\
    \includegraphics[width=\linewidth, trim={0 1330 0 690}, clip]{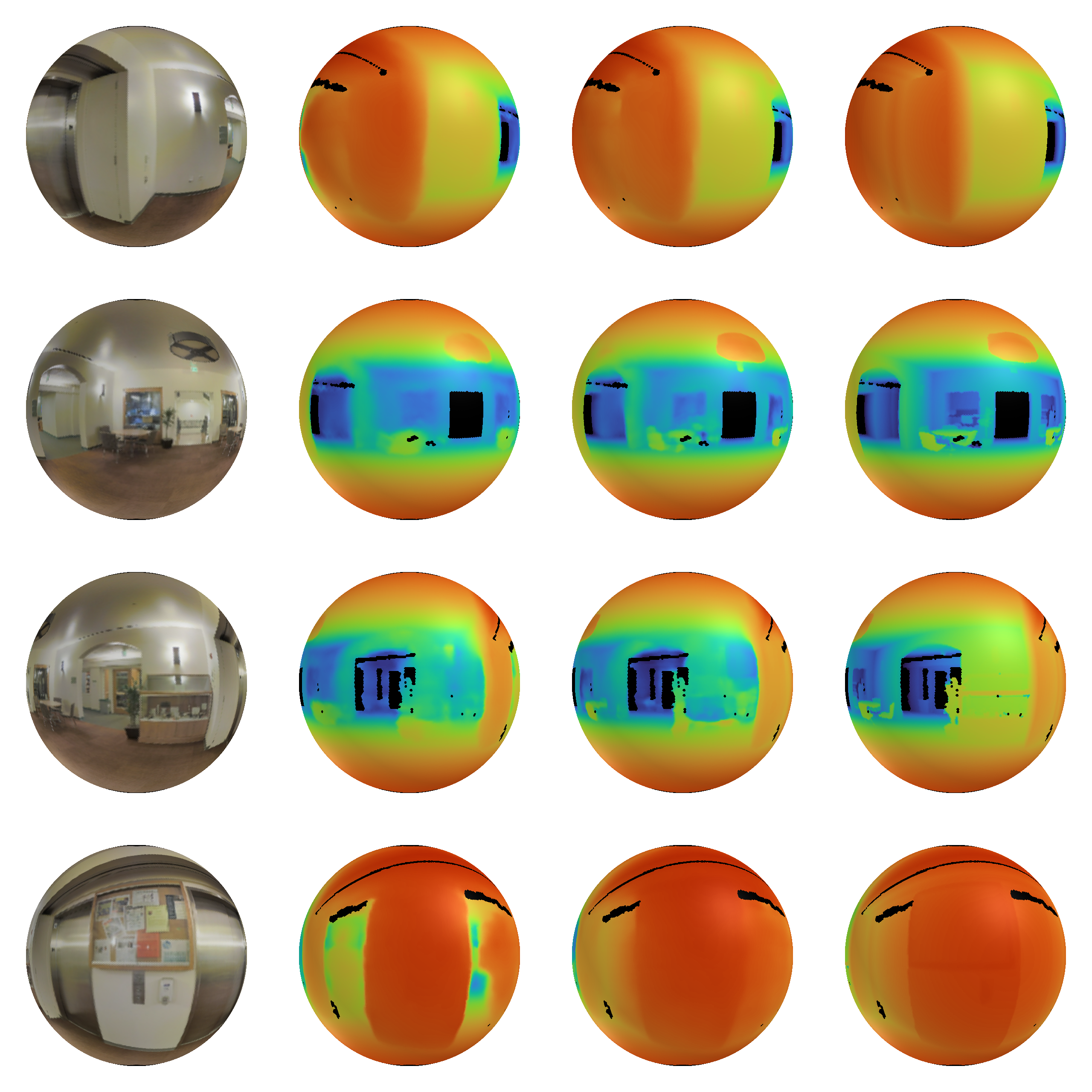} \\
    \begin{tabular}{@{}c@{}c@{}c@{}c@{}}
        \includegraphics[width=0.25\linewidth, trim={0 1970 1920 50}, clip]{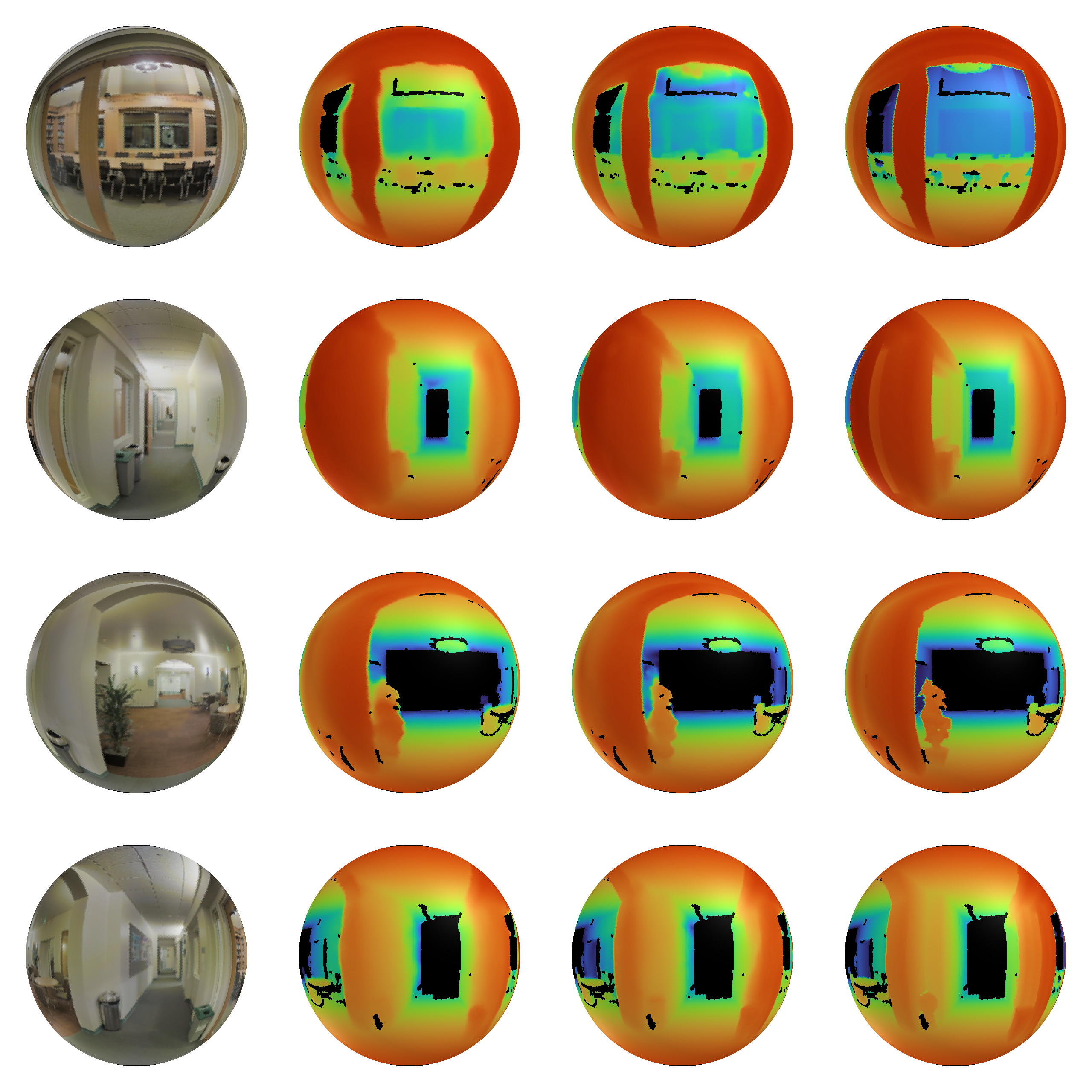} &
        \includegraphics[width=0.25\linewidth, trim={640 1970 1280 50}, clip]{images/supp/high_res/0033.png} &
        \includegraphics[width=0.25\linewidth, trim={1280 1970 640 50}, clip]{images/supp/high_res/0033.png} &
        \includegraphics[width=0.25\linewidth, trim={1920 1970 0 50}, clip]{images/supp/high_res/0033.png} \\
        RGB & Rank 7 & Rank 8 & GT \\
    \end{tabular}
    
    \caption{\textbf{High resolution results.} \textbf{From left to right:} RGB, Rank 7, Rank 8, GT. Results in rank 8 are sharper and visibly more accurate in their depth estimation.}
    \label{supp:fig:high_res_vis}
\end{figure*}

\subsection{Qualitative High-Resolution Comparison}
In \cref{supp:fig:high_res_vis}, we show the depth estimation results of rank 7 and rank 8 models. As specified in \cref{supp:protocol/high_res}, the rank 8 model was finetuned on rank 7 pretrained weights. Beyond the better quantitative performance shown in \cref{tab:results_1024} (in the main paper), these results show that the higher resolution model produces a sharper and more accurate depth image.

\putbib
\end{bibunit}

\end{document}